\documentclass[conference]{IEEEtran}
\usepackage[letterspace=-90]{microtype}
\usepackage{amsmath,amssymb,amsfonts,enumerate,multirow,xfrac,url,bm}
\usepackage{algorithmic,lipsum}
\usepackage{graphicx}
\usepackage{textcomp}
\usepackage{xcolor}
\usepackage{upgreek}
\usepackage{manfnt,yfonts}
\usepackage{algorithm,algorithmic}
\usepackage[caption=false, font=footnotesize]{subfig}
\graphicspath{{./images/}}

\usepackage{definice}
\usepackage{mathrsfs}

\usepackage{float}

\newcommand{\imageHeight}{\textls{\texttt{image\_\,height}}}
\newcommand{\imageWidth}{\textls{\texttt{image\_\,width}}}
\newcommand{\framerate}{\textls{\texttt{frame\_\,rate}}}

\newcommand{\T}{^{\top}}
\newcommand{\Expect}[1]{\mathrm{E}\!\left(\, #1\, \right)}
\newcommand{\cov}[1]{\text{cov}\!\left(\, #1\, \right)}
\newcommand{\var}[1]{\text{var}\!\left(\, #1\, \right)}

\newcommand{\RMSE}{ \mathrm{RMSE} }
\newcommand{\ANEES}{ \mathrm{ANEES} }

\newcommand{\ms}{\hspace{0.17cm}} 
\newcommand{\px}{\texttt{|\hspace{-0.03cm}p\hspace{-0.03cm}x\hspace{-0.03cm}|}}

\newcommand{\twoDee}{ \mathrm{2 \hspace{-0.03cm} D} }
\newcommand{\threeDee}{ \mathrm{3 \hspace{-0.03cm} D} }
\newcommand{\meanScalar}[1]{m_{#1}}

\newcommand{\xTwoDee}{\mathsf{x}}
\newcommand{\xVelocityTwoDee}{\dot{\mathsf{x}}}
\newcommand{\yTwoDee}{\mathsf{y}}
\newcommand{\yVelocityTwoDee}{\dot{\mathsf{y}}}
\newcommand{\widthTwoDee}{\upomega}
\newcommand{\widthVelocityTwoDee}{\dot{\upomega}}
\newcommand{\heightTwoDee}{\mathsf{h}}
\newcommand{\heightVelocityTwoDee}{\dot{\mathsf{h}}}
\newcommand{\xThreeDee}{x}
\newcommand{\xVelocityThreeDee}{\dot{x}}
\newcommand{\yThreeDee}{y}
\newcommand{\yVelocityThreeDee}{\dot{y}}
\newcommand{\zThreeDee}{z}
\newcommand{\zVelocityThreeDee}{\dot{z}}
\newcommand{\widthThreeDee}{\omega}
\newcommand{\heightThreeDee}{h}

\newcommand{\LOneScale}{0.6}\newcommand{\LOneRise}{0.06cm}
\newcommand{\annotationLGD}{(\raisebox{\LOneRise}{\includegraphics[scale=\LOneScale]{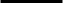}})}
\newcommand{\detectionsLGD}{(\raisebox{0.04cm}{\includegraphics[scale=0.25]{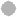}})}
\newcommand{\twooDeeLGD}{(\raisebox{\LOneRise}{\includegraphics[scale=\LOneScale]{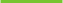}})}
\newcommand{\BoTSORTLGD}{(\raisebox{\LOneRise}{\includegraphics[scale=\LOneScale]{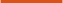}})}
\newcommand{\proposedLGD}{(\raisebox{\LOneRise}{\includegraphics[scale=\LOneScale]{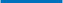}})}
\newcommand{\camBoundariesLGD}{(\raisebox{\LOneRise}{\includegraphics[scale=\LOneScale]{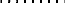}})}

\newcommand{\twooDeeStar}{(\raisebox{0.02cm}{\includegraphics[scale=0.6]{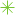}})}
\newcommand{\BoTSORTStar}{(\raisebox{0.02cm}{\includegraphics[scale=0.6]{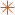}})}
\newcommand{\proposedStar}{(\raisebox{0.02cm}{\includegraphics[scale=0.6]{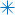}})}

\newcommand{\BoT}{\mathrm{B \hspace{-0.03cm} o \hspace{-0.03cm} T}}

\newcommand{\Fncv}{\mathbf{F}_{\text{\textls{NCV}}}}

\newcommand{\blkdiag}[1]{\text{diag} \hspace{-0.05cm} \left(\, {#1} \, \right)}

\newcommand{\dummyVar}{\mathtt{p}}
\newcommand{\dummyVel}{\mathtt{v}}

\begin{document}

\onecolumn
\vspace*{5cm}
This paper has been accepted for publication in 2024 27th International Conference on Information Fusion (FUSION)
Please cite the paper as:
J. Krejčí, O. Kost, O. Straka and J. Duník, "Pedestrian Tracking with Monocular Camera using Unconstrained 3D Motion Model," 2024 27th International Conference on Information Fusion (FUSION), Venice, Italy, 2024, pp. 1-8, doi: 10.23919/FUSION59988.2024.10706432.
\clearpage
\twocolumn

\IEEEoverridecommandlockouts
\title{
    Pedestrian Tracking with Monocular Camera using Unconstrained 3D Motion Model
	\thanks{
        This research was co-funded by the European Union under the project ROBOPROX (reg. no. CZ.02.01.01/00/22\_008/0004590) and Technology Agency of the Czech Republic, programme National Competence Centres, proj.~\# TN 0200 0054 Bozek Vehicle Engineering National Competence Center. 
	}
}

\author{
 \IEEEauthorblockN{Jan Krejčí, Oliver Kost, Ondřej Straka, Jindřich Duník}
 \IEEEauthorblockA{
\textit{Department of Cybernetics, University
  of West Bohemia}, Pilsen, Czech Republic
 }
 \IEEEauthorblockA{
 Email: \{ jkrejci, kost, straka30, dunikj \}@kky.zcu.cz
 }
 }

\maketitle

\begin{abstract}
    A first-principle single-object model is proposed for pedestrian tracking.
    It is assumed that the extent of the moving object can be described via known statistics in 3D, such as pedestrian height. 
    The proposed model thus need not constrain the object motion in 3D to a common ground plane, which is usual in 3D visual tracking applications. 
    A nonlinear filter for this model is implemented using the unscented Kalman filter (UKF) and tested using the publicly available MOT-17 dataset.
    The proposed solution yields promising results in 3D while maintaining excellent results when projected into the 2D image.
    Moreover, the estimation error covariance matches the true one.
	Unlike conventional methods, the introduced model parameters have convenient meaning and can readily be adjusted for a problem.
\end{abstract}

\begin{IEEEkeywords}
	Visual object tracking, bounding box, unscented Kalman filter, 3D modeling
\end{IEEEkeywords}

\section{Introduction}
Tracking algorithms play a crucial role in many domains of today's society.
Indeed, accurate information about the position of nearby objects is essential for safe and effective land, air, and marine transportation~\cite{Bar-Shalom-et.al:2011}.
The tracking algorithms are thus a backbone of any surveillance system that monitors e.g., pedestrians in the vicinity of vehicles.
These algorithms typically belong to the class of \emph{visual tracking systems} (VTSs) processing camera readings~\cite{SonkaHlavacBoyle:ImageProcessing:2008}.
This paper focuses on the tracking-by-detection paradigm, where a \emph{visual detection network} (VDN) pre-processes the image data for the tracker.

Traditional monocular VTS represent the tracked object using a bounding box in 2D space defined by the camera image~\cite{SORT:2016,BoT-SORT:2022,KrKoStDu:2023_FUSION,KrKoSt:2023_FUSION}.
Estimating object 3D position from a monocular camera is inherently an ill-posed problem due to the impossibility of determining the distance between the object and the camera. 
This information is, however, \textit{essential} for path planning \cite{Surface3DMonoCamTracking:2022} or occlusion handling~\cite{MonocularMTT3dOcclu:2013}, to name a few.
To obtain depth information, we have to either employ another sensor \emph{or} introduce additional knowledge/assumption on the tracked objects.
The \textit{former} option includes utilization of a stereo camera~\cite{UKF3D_StereoCam:2012}, LiDAR~\cite{KITTI-dataset:2012}, 
or custom VDN providing, e.g., depth measurements~\cite{Mono-Camera3D_PMBM:2018}, which, in the end, increases the cost of the VTS and possibly the computational demands of the algorithms.
The \textit{latter} option constrains the motion of the objects to a common ground plane, which, however, requires the ground plane to be given~\cite{MonocularMTT3dOcclu:2013,GroundPlaneUKF3d:2008} or estimated~\cite{Surface3DMonoCamTracking:2022}, \cite{GroundPlaneEstim:2017}. 

\begin{figure}
\vspace{-2mm}
	\centering
	\subfloat[Filtering results in 2D.\label{fig:front2D}]{ \includegraphics[width=0.4\linewidth]{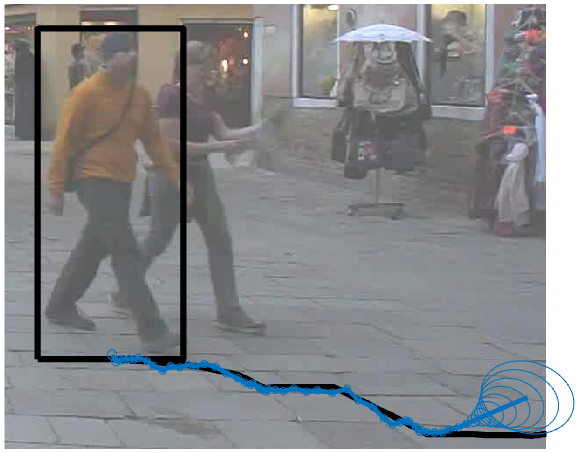} }
	\subfloat[Filtering results in 3D, $\xThreeDee$-$\zThreeDee$ view.\label{fig:front3D}]{ \includegraphics[width=0.52\linewidth]{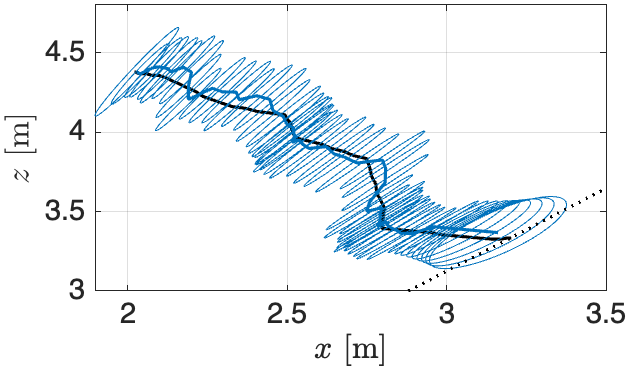} }
 \vspace{-1mm}
	\caption{Results of the proposed filter including error covariance ellipses \proposedLGD{} compared to annotations \annotationLGD{}. Faster R-CNN detections were used, and they were not available by the end of the scenario. Notice that the uncertainty is largest in the direction of the line of sight.}\label{fig:front}
\vspace{-6mm}
\end{figure}

This paper develops a linear stochastic dynamic model and the corresponding nonlinear measurement model suitable for the monocular pedestrian VTS, that \textit{(i)} allows unconstrained object motion description in 3D space and \textit{(ii)} does not rely on the common ground plane assumption.
Instead, the proposed BB dynamics model is designed under the assumption of a pedestrian with average body proportions.
This idea shares some similarities with \cite{MonocularMTT3dOcclu:2013,PedestrianKF_NCV:1997,KFpedestrianStereo_WH:2004}.
In addition, techniques for setting and identifying the resulting state-space model parameters are discussed and illustrated using available datasets.
A numerically stable \emph{unscented Kalman filter} (UKF) is designed for the proposed model and is analyzed using the MOT-17 dataset~\cite{MOT-16:2016,MOT17-webpage:2023}.
The results are illustrated in Fig.~\ref{fig:front}.

The paper is organized as follows.
Section~II provides background for monocular VTSs.
Traditional tracking algorithms working in 2D are reviewed in Section~III and
the proposed method is presented in Section~IV.
A practical comparison is given in Section~V, and the paper concludes in Section~VI.

\section{Formalization of the Problem}
\vspace{-1mm}
This section outlines the relation between a moving object in 3D and its observation via a VDN applied to a camera image.
The motion in 3D is discussed, followed by the projection of the object at a particular time into a pinhole camera image.
Relation of the projection to the VDN output is described last.

\subsection{Object Motion in 3D}
\vspace{-1mm}
A moving object can be comprehensively described using the \emph{3D state} vector $\mathbf{x}^{\threeDee}(t)\in\mathbb{R}^{n_{ \mathbf{x} }^{ \threeDee } }$ at any time $t\in\mathbb{R}$, containing variables such as a chosen reference point, velocity, extent, or shape of the object.
According to the laws of physics, these variables would naturally change in \emph{continuous} time. 
Here, the variable dynamics is described in discrete time by 
\begin{align}
	\mathbf{x}_{k+1}^{\threeDee} = \mathbf{f}_k( \mathbf{x}_k^{\threeDee}, \mathbf{w}_k^{\threeDee}), \label{eq:dynamics_nonlinear_general}
\end{align}
where $\mathbf{x}_k^{\threeDee}=\mathbf{x}^{\threeDee}(t_k)$ for some time-instant $t_k=T\!\cdot\! k$, where $T$ is the sampling period and $k\in\mathbb{N}$.
The vector $\mathbf{w}_k^{\threeDee}\in\mathbb{R}^{n_{\mathbf{w}}^{\threeDee} }$ is unknown noise vector,
and $\mathbf{f}_k$ describes the dynamics of $\mathbf{x}_k^{\threeDee}$.
Although generally unknown, Eq.~\eqref{eq:dynamics_nonlinear_general} can be described using standard physical models rooted in continuous time~\cite{Bar-Shalom-et.al:2011}.

\subsection{Pinhole Camera Model}\label{sec:CameraModel}
\begin{figure}
	\centering
	\subfloat[Projection to focal plane.\label{fig:cam:a}]{ \includegraphics[width=0.35\linewidth]{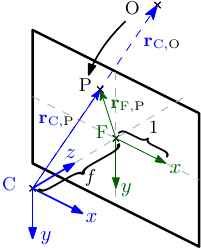} }
	\subfloat[Focal and image coordinates.\label{fig:cam:b}]{ \includegraphics[width=0.45\linewidth]{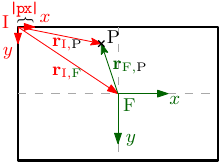} }
	\caption{Illustration of the geometric transformations for a point $\mathrm{O}$ in 3D. Camera coordinates are denoted with blue, the focal plane with green, and the image plane with red. Units of measurement are meters.}\label{fig:camera_geometry_1}
\vspace{-6mm} 
\end{figure}

The pinhole camera model describes the relationship between an object in 3D and its projection to the idealized pinhole camera image, i.e., the \emph{perspective projection}~\cite{SonkaHlavacBoyle:ImageProcessing:2008,PuttingObjectsInPerspective:2008}.
The corresponding geometry is illustrated in Fig.~\ref{fig:camera_geometry_1}.
By not explicitly considering camera motion in this paper, it is assumed that the world coordinates are equal to the \emph{camera} coordinates\footnote{
    Note that this assumption can be relaxed once the position and orientation of the camera coordinates w.r.t.~the world coordinates are known.
} denoted with~$\mathrm{C}$, being expressed in meters.
The resulting \emph{image} $\mathrm{I}$ is assumed to be composed of square pixels 
with side length denoted \px{} in meters.
The perspective projection of an object in 3D can then be described by first projecting it to the \emph{focal} plane $\mathrm{F}$ being $f$ meters in front of the camera center and then describing the result in the image plane $\mathrm{I}$.
\subsubsection{Projection of a 3D Point into Focal Plane}
Denote the vector describing a 3D point $\mathrm{O}$ of the \emph{object} in the camera coordinates $\mathrm{C}$ with $\mathbf{r}_{\mathrm{C} , \mathrm{O}} = [\xThreeDee\ \yThreeDee\ \zThreeDee]$.
The projection of $\mathrm{O}$ to the focal plane $\mathrm{F}$, denoted with $\mathrm{P}$ described by the vector $\mathbf{r}_{\mathrm{F}, \mathrm{P}}$ is 
\begin{align}
	 \mathbf{r}_{\mathrm{F} , \mathrm{P}} = \tfrac{f}{ \zThreeDee }
            \left[\begin{smallmatrix}
	 			1 & 0 & 0 \\
	 			0 & 1 & 0
            \end{smallmatrix}\right]
	 		\mathbf{r}_{\mathrm{C} , \mathrm{O}}. \label{eq:3Dto2dm}
\end{align}

If a velocity $\dot{\mathbf{r}}_{ \mathrm{C} , \mathrm{O}} = [\xVelocityThreeDee \ \yVelocityThreeDee\ \zVelocityThreeDee]$ is associated with $\mathrm{O}$, its projection $\dot{\mathbf{r}}_{ \mathrm{F} , \mathrm{P}}$ to $\mathrm{F}$ is given by differentiating \eqref{eq:3Dto2dm} w.r.t. time,
\begin{align}
	\dot{\mathbf{r}}_{\mathrm{F} , \mathrm{P}} = \tfrac{f}{ \zThreeDee }
    \left[\begin{smallmatrix}
		1 & 0 & 0 \\
		0 & 1 & 0
    \end{smallmatrix}\right]
	\left( \dot{\mathbf{r}}_{\mathrm{C} , \mathrm{O}} - \tfrac{ \zVelocityThreeDee }{ \zThreeDee } \mathbf{r}_{\mathrm{C} , \mathrm{O}} \right). \label{eq:3Dto2dm:velocity}
\end{align}

\begin{figure}
	\centering
	\includegraphics[width=0.8\linewidth]{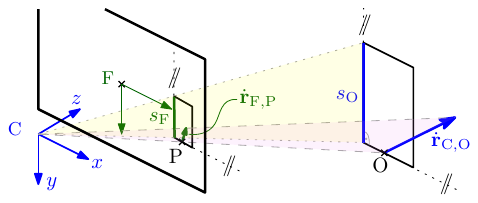}
	\caption{Illustration of the geometric transformation of a line segment (a height) $s_{\mathrm{O}}$ and a velocity $\dot{\mathbf{r}}_{\mathrm{C,O}}$ in 3D to the corresponding variables $s_{\mathrm{F}}$ and $\dot{\mathbf{r}}_{\mathrm{F,P}}$ in the focal plane.}\label{fig:camera_geometry_2}
\vspace{-6mm} 
\end{figure}

Consider a line segment in 3D such that \emph{(i)} its both end points have the same depth $\zThreeDee$ and velocity in depth $\zVelocityThreeDee$, and \emph{(ii)} it is aligned either with the horizontal or vertical axis, such as the width or height of a perfectly aligned rectangle with the focal plane as illustrated in Fig.~\ref{fig:camera_geometry_2}.
Denoting the length of the line segment in meters in 3D and its temporal change rate with $s_{ \mathrm{O} }$ and $\dot{s}_{ \mathrm{O} }$, respectively, the projections of these variables into the focal plane $\mathrm{F}$ are given by 
\begin{align}
	s_{ \mathrm{F} } &= \tfrac{f}{z} s_{ \mathrm{O} }, &
	\dot{s}_{ \mathrm{F} } &= \tfrac{f}{ \zThreeDee } \left( \dot{s}_{ \mathrm{O} } - \tfrac{\zVelocityThreeDee }{ \zThreeDee } s_{ \mathrm{O} } \right). \label{eq:lineSegment:3DtoFocal}
\end{align}

\subsubsection{Basis Change to Image Plane}
The transformation of the point $\mathrm{P}$ described in the focal plane $\mathrm{F}$ denoted by the vector $\mathbf{r}_{\mathrm{F} , \mathrm{P} }$ to the image plane $\mathrm{I}$ is given by
\begin{align}
	\mathbf{r}_{\mathrm{I} , \mathrm{P}} =
	\tfrac{1}{\px}\!\cdot\!
	\mathbf{r}_{\mathrm{F} , \mathrm{P}} + \mathbf{r}_{\mathrm{I} , \mathrm{F}},
\end{align}
where $\mathbf{r}_{\mathrm{I}, \mathrm{F}}$ expressed in pixels is the vector pointing from the origin of the image plane $\mathrm{I}$ to the center of the projection $\mathrm{F}$ as illustrated in Fig.~\ref{fig:cam:b}.
The velocity $\dot{\mathbf{r}}_{\mathrm{F} , \mathrm{P}}$, the length $s_{ \mathrm{F} }$ and its temporal change rate $\dot{s}_{ \mathrm{F} }$ are transformed from $\mathrm{F}$ to $\mathrm{I}$ by
\begin{align}
	\dot{\mathbf{r}}_{\mathrm{I} , \mathrm{P}} &= 
	\tfrac{1}{\px}\!\cdot\!
		\dot{\mathbf{r}}_{\mathrm{F} , \mathrm{P}}, &
	s_{ \mathrm{I} } &= \tfrac{1}{\px}\!\cdot\! s_{ \mathrm{F} }, & \dot{s}_{ \mathrm{I} } &= \tfrac{1}{\px}\!\cdot\! \dot{s}_{ \mathrm{F} }. \label{eq:FocalToImage:others}
\end{align}.

The intrinsic camera parameters $\px$ and $f$ describing the projection are assumed to be known (e.g., calibrated~\cite{SonkaHlavacBoyle:ImageProcessing:2008}).
\\[-2mm]

Assuming the above projection rules suffice to transform all elements of $\mathbf{x}_k^{\threeDee}$, one can define a vector $\mathbf{x}_k^{\twoDee} \in \mathbb{R}^{n_{ \mathbf{x} }^{ \twoDee }}$ describing the object in 2D, by stacking the projected variables.
Although the \emph{2D state} $\mathbf{x}_k^{\twoDee}$ may contain velocities, it cannot describe the original object comprehensively as information is being lost by discarding the $z$-vector entries in~\eqref{eq:3Dto2dm} and~\eqref{eq:3Dto2dm:velocity}.

To distinguish between variables in 3D in meters and their 2D projections in pixels, the latter will be denoted with math sans-serif, e.g., the position coordinate $\xTwoDee$ denotes the projection of $\xThreeDee$. 
The following subsection introduces a conventional description of the object by discarding even more information from~$\mathbf{x}_k^{\twoDee}$.

\subsection{Object's Bounding Box and Annotation}
A handy description of the object in the image plane is provided by overlaying it with a \emph{bounding-box} (BB), i.e., a rectangle, referred to as \emph{object's BB} and denoted with $\mathbf{b}_k\in \mathbb{R}^4$ in pixels.
A linear relation is assumed to exist, such as
\begin{align}
    \mathbf{b}_k = \mathbf{H} \mathbf{x}_k^{\twoDee} = [\xTwoDee_k\ \yTwoDee_k\ \widthTwoDee_k\ \heightTwoDee_k]\T \in\mathbb{R}^4, \label{eq:annotation}
\end{align}
where $\mathbf{H}\in\mathbb{R}^{ 4 \times n_{\mathbf{x}}^{\twoDee} }$ is known, $\xTwoDee_k$ and $\yTwoDee_k$ are the position coordinates e.g., of lower-bottom center of the projected object, while $\widthTwoDee_k$ and $\heightTwoDee_k$ are its width and height, respectively.

Such a description is usually simple enough to be obtained by hand from an image, which is referred to as \emph{annotation}, denoted with $\mathbf{a}_k$.
The annotations acquisition is a time-consuming and costly process; therefore, they are usually unavailable.

\subsection{Bounding Box Detection Modeling}
A VDN produces BBs separately for each video frame.
Some VDNs provide additional information such as class likelihoods, visual features, etc.
Such information is not considered in this paper. 
Each output of a VDN for a frame $k$ is a detection $\mathbf{z}_k$ forming a BB with its position and extent, such as in~\eqref{eq:annotation}.

Although VDNs are designed to bound desired objects, the resulting detections are not guaranteed to do so.
Some objects may be miss-detected, multiple detections might be related to the same object, or the VDN may produce false detections.
Since this paper does \emph{not} focus on data association techniques, the detections are a~priori assumed to be associated with an object.
For this purpose, technique from~\cite{KrKoSt:2023_FUSION} is used here.
As a result, it is assumed that for an object's BB (hereafter simply referred to as BB) $\mathbf{b}_k$ at time step $k=0,1,\dots,K$, there is either no detection or exactly one detection $\mathbf{z}_k$ in which case
\begin{align}
	\mathbf{z}_k = \mathbf{Hx}_k^{\twoDee}  + \mathbf{v}_k, \label{eq:detection}
\end{align}
where $\mathbf{v}_k$ is the measurement noise.
The complexity of the $\mathbf{v}_k$ description depends on the VDN properties, see~\cite{KrKoSt:2023_FUSION}.
\\

Observe that the object's depth is not measured directly by the VDN.
Therefore, the estimation of the 3D state may become ill-conditioned.
Unless a common ground plane is assumed to be known, current estimation algorithms work in 2D to overcome this and make limited use of the above principles.

\section{Baseline Filtering Techniques in 2D}
This section presents two baseline algorithms.
The first one is constructed based on the previous work~\cite{KrKoStDu:2023_FUSION,KrKoSt:2023_FUSION}, aimed at careful motion and measurement modeling. 
The second algorithm insinuates a typical approach in VTS literature and is taken from the state-of-the-art \emph{"bag of tricks" for simple, online and real-time tracking} (BoT-SORT) VTS~\cite{BoT-SORT:2022}.

\subsection{2D-Model-Based Linear Time-Invariant Filter}\label{sec:twoDeeFilter}
Papers \cite{KrKoStDu:2023_FUSION} and \cite{KrKoSt:2023_FUSION} present linear time-invariant motion\footnote{
	The specific model used in this paper is referred to as \emph{BB dynamic model~2} in~\cite{KrKoStDu:2023_FUSION}, and the notation used herein is slightly varied.
} and measurement models suitable for VTSs, including identification of their parameters.
The motion model from~\cite{KrKoStDu:2023_FUSION} uses the \emph{nearly constant velocity} (NCV) model, which is described first.
Using it as a building block, the state-space model is introduced, followed by the construction of a filter for this model.
Note that this resulting filter has not yet been presented elsewhere.

\subsubsection{Nearly Constant Velocity Model}\label{sec:NCV}
Consider a stochastic process $\dummyVar$ exhibiting significant time correlation.
The NCV model assumes that the time derivative of the process $\dummyVar$ can be sufficiently described by a random process $\dummyVel$ driven by a white noise, which can be heuristically written as
\begin{align}
	\tfrac{d\dummyVar(t)}{dt}&=\dummyVel(t), & \tfrac{d\dummyVel(t)}{dt} &= w(t),\ \var{w(t)}= q\delta(0), \label{ncv_continuous_assumption}
\end{align}
where $q>0$ is the \emph{power spectral density} (PSD) of the zero-mean continuous white noise $w(t)$, and $\delta(t)$ is the Dirac delta function\footnote{A correct mathematical notation would involve a stochastic differential equation and a Wiener process both of which are not needed in this paper \cite{Bar-Shalom-et.al:2011}.}.
Its discretization leads to \cite{Bar-Shalom-et.al:2011} 
\vspace{-0mm}
\begin{align}
    \left[\begin{smallmatrix}   
		\dummyVar_{k+1} \\ \dummyVel_{k+1}
    \end{smallmatrix}\right]
    = \underbrace{
        \left[\begin{smallmatrix}
    			1 & T \\ 0 & 1
        \end{smallmatrix}\right]
    }_{\Fncv} 
    \left[\begin{smallmatrix}
		\dummyVar_{k} \\ \dummyVel_k
    \end{smallmatrix}\right]
	\!+\!
	\mathbf{w}_k,\
	\cov{ \bfw_k } = 
	q \! 
	\underbrace{
        \left[\begin{smallmatrix}
			\sfrac{T^3}{3} & \sfrac{T^2}{2} \\
			\sfrac{T^2}{2} & T
        \end{smallmatrix}\right]
	}_{ \mathbf{T} }
	\!,\!
	\label{eq:NCV}
\end{align}
\\[-3mm]
with the state noise covariance being a function of the sampling period $T=\sfrac{1}{\framerate}$ in seconds and a parameter $q$ in m$^2$s$^{-3}$.

\subsubsection{State-Space Modeling Based on~\cite{KrKoStDu:2023_FUSION,KrKoSt:2023_FUSION}}
In the notation of this paper,
the state defined in \cite{KrKoStDu:2023_FUSION} describes the moving object in 2D with
\begin{align}
	\mathbf{x}_k^{\twoDee} = [\xTwoDee_k\ms \xVelocityTwoDee_k\ms \yTwoDee_k\ms \yVelocityTwoDee_k\ms \widthTwoDee_k\ms \widthVelocityTwoDee_k\ms \heightTwoDee_k\ms \heightVelocityTwoDee_k]\T, \label{eq:2Dstate:variables}
\end{align}
where $\xTwoDee_k, \yTwoDee_k, \widthTwoDee_k$ and $\heightTwoDee_k$ have the same meaning as in~{\eqref{eq:annotation}}, while the variables $\xVelocityTwoDee_k, \yVelocityTwoDee_k, \widthVelocityTwoDee_k$ and $\heightVelocityTwoDee_k$ represent their respective temporal change rates, i.e., velocities.
For the motion model, the pairs $\xTwoDee_k, \xVelocityTwoDee_k;\, \yTwoDee_k, \yVelocityTwoDee_k;\, \widthTwoDee_k, \widthVelocityTwoDee_k;\, $ and $\heightTwoDee_k, \heightVelocityTwoDee_k$ are each assumed to obey independent NCV model~\eqref{eq:NCV}, leading to 
\begin{subequations}\label{eqs:linear-system}
	\begin{align}
		\mathbf{x}_{k+1}^{\twoDee} &= \mathbf{F}^{\twoDee} \mathbf{x}_{k}^{\twoDee} + \mathbf{w}_{k}^{\twoDee}, \label{eq:system:motion} \\
		\mathbf{z}_k &= \mathbf{H} \mathbf{x}_k^{\twoDee} + \mathbf{v}_k,	\label{eq:system:measurement}
	\\[-6mm]\nonumber
        \end{align}
\end{subequations}
where the dynamic and measurement matrices are given by,
\begin{align}
	\mathbf{F}^{\twoDee} &= \blkdiag{ \mathbf{F}_{\mathrm{NCV}}, \mathbf{F}_{\mathrm{NCV}}, \mathbf{F}_{\mathrm{NCV}}, \mathbf{F}_{\mathrm{NCV}}}, \\
		\mathbf{H} &= 
	\left[\begin{smallmatrix}
		1 & 0 & 0 & 0 & 0 & 0 & 0 & 0 \\
		0 & 0 & 1 & 0 & 0 & 0 & 0 & 0 \\
		0 & 0 & 0 & 0 & 1 & 0 & 0 & 0 \\
		0 & 0 & 0 & 0 & 0 & 0 & 1 & 0
	\end{smallmatrix}\right],\label{eq:measurement-matrix-numeric}
\end{align}
respectively, where $\blkdiag{\cdot}$ denotes a block diagonal matrix.
For~\eqref{eq:system:measurement} to be well-defined, even columns of the measurement matrix $\mathbf{H}$~\eqref{eq:measurement-matrix-numeric} are in seconds, and the rest are unitless.
The noises $\mathbf{w}_k^{\twoDee}$ and $\mathbf{v}_k$ are assumed to be zero mean and uncorrelated, and uncorrelated with the state. 
Given the NCV models, the state noise covariance matrix is then
\begin{align}
    \mathbf{Q}^{\twoDee} = \cov{ \mathbf{w}_k^{\twoDee} } = \gamma^2 \cdot \blkdiag{ q_{ \xVelocityTwoDee } \mathbf{T}, q_{ \yVelocityTwoDee } \mathbf{T}, q_{ \widthVelocityTwoDee } \mathbf{T}, q_{ \heightVelocityTwoDee } \mathbf{T} },
\end{align}
where the parameters\footnote{
    Alternative setup of the PSDs $q_{\xVelocityTwoDee}$ and $q_{\yVelocityTwoDee}$, and several measurement noise covariances was discussed in \cite{PedestrianKF_NCV:1997}.
}
$q_{\xVelocityTwoDee}\! =\! 0.011, q_{\yVelocityTwoDee}\! =\! 0.037, q_{\widthVelocityTwoDee}\! =\! 0.013$ and $q_{\heightVelocityTwoDee}\! =\! 0.025$
were identified in~\cite{KrKoStDu:2023_FUSION} from the MOT datasets~\cite{MOT-15:2015,MOT17-webpage:2023,MOT-20:2020}, and the measurement noise covariance matrix is 
\begin{align}
    \mathbf{R} \!=\! \cov{ \mathbf{v}_k } \!= \gamma^2
    \! \cdot \! 10^{-5} \! \left[\begin{smallmatrix}
       \ms2.232 &\ms0.086 & -0.787  &  -0.084 \\
       \ms0.086 &\ms2.817 &\ms0.080 &  -2.280 \\
        -0.787  &\ms0.080 &\ms2.036 &\ms0.266 \\
        -0.084  & -2.280  &\ms0.266 &\ms4.661 
    \end{smallmatrix}\right]\!\!,
    \label{eq:state-noise-covariance-matrix:identification}
\end{align}
where $\gamma=\min( \imageWidth, \imageHeight )$ was identified\footnote{
	The matrix $\mathbf{R}$ is denoted in~\cite{KrKoSt:2023_FUSION} as $\mathbf{R}_{\mathrm{U}}$.
} in~\cite{KrKoSt:2023_FUSION} for the Faster R-CNN detector applied to the MOT-17 dataset~\cite{MOT-16:2016,MOT17-webpage:2023}.
The parameters were identified for pedestrians, but the techniques from \cite{KrKoStDu:2023_FUSION,KrKoSt:2023_FUSION} can also be used for different objects, such as vehicles.

\subsubsection{2D-Model-Based Filter for~\eqref{eqs:linear-system}}
An optimal unbiased state estimator having minimal \emph{mean squared error} (MSE) for the linear system such as~\eqref{eqs:linear-system} can be derived in a closed form.
It is the Kalman filter (KF), formed by a recursion composed of the prediction and filtering steps.

The recursion starts at $k=0$ setting the initial filtering estimate $\mathbf{x}_{0|0}^{\twoDee} = \Expect{ \mathbf{x}_0^{\twoDee} \,|\, \mathbf{z}_0 } $ and its error covariance matrix $ \mathbf{P}_{0|0}^{\twoDee} = \cov{ \mathbf{x}_0^{\twoDee} \,|\, \mathbf{z}_0 }$.
The subscript $k|l$ means that the corresponding estimate at the time step $k$ is conditioned on measurements up to the time step $l$.
For $k=1,2,\dots,K$, the prediction step is given by
\begin{subequations}
	\begin{align}
 \nonumber\\[-6mm]
		\mathbf{x}_{k|k\!-\!1}^{\twoDee} &= \mathbf{F}^{\twoDee} \mathbf{x}_{k\!-\!1|k\!-\!1}^{\twoDee}, \\
		\mathbf{P}_{k|k\!-\!1}^{\twoDee} &= \mathbf{F}^{\twoDee} \mathbf{P}_{k\!-\!1|k\!-\!1}^{\twoDee} \left(\mathbf{F}^{\twoDee}\right)\T + \mathbf{Q}^{\twoDee},
	\\[-8mm]\nonumber
        \end{align}
\end{subequations}
and the filtering step is given by
\begin{subequations}\label{eqs:KF:filter}
	\begin{align}
		\!\!\!\!\! \mathbf{x}_{k|k}^{\twoDee} \! &= \! \mathbf{x}_{k|k\!-\!1}^{\twoDee} + \mathbf{K}_k^{\twoDee} ( \mathbf{z}_k \! - \! \mathbf{H}\mathbf{x}_{k|k\!-\!1}^{\twoDee} ), \label{eq:KF_update_mean} \\
		\!\!\!\!\! \mathbf{P}_{k|k}^{\twoDee} \! &= \! (\mathbf{I} \! - \! \mathbf{K}_k^{\twoDee} \mathbf{H}) \mathbf{P}_{k|k\!-\!1}^{\twoDee} \! (\mathbf{I} \! - \! \mathbf{K}_k^{\twoDee} \mathbf{H})\T \!\! + \! \mathbf{K}_k^{\twoDee} \mathbf{R} \! \left( \mathbf{K}_k^{\twoDee} \right)\T \!\!\!\!, \! \\
		\!\!\!\!\! \mathbf{K}_k^{\twoDee} \! &= \! \mathbf{P}_{k|k\!-\!1}^{\twoDee} \mathbf{H}\T ( \mathbf{H} \mathbf{P}_{k|k\!-\!1}^{\twoDee} \mathbf{H}\T + \mathbf{R} )^{-1},
	\end{align}
\end{subequations}
where $\mathbf{K}_k^{\twoDee}$ is the Kalman gain.
The error corresponding to the predictive and filtering estimates $\mathbf{x}_{k|k\!-\!1}^{\twoDee}$ and $\mathbf{x}_{k|k}^{\twoDee}$, respectively, are characterized by the predictive and filtering covariance matrices $\mathbf{P}_{k|k\!-\!1}^{\twoDee}$ and $\mathbf{P}_{k|k}^{\twoDee}$, respectively.
The filtering step~\eqref{eqs:KF:filter} is performed only when a measurement $\mathbf{z}_k$ exists at the time step $k$, which generally applies throughout the paper.

\subsubsection{Filter Initialization}\label{sec:BBmod2:initialization}
For a fair comparison, the filter is initialized using the measurement $\mathbf{z}_0$ as\footnote{
    The matrix $\mathbf{H}$ appearing in~\eqref{eq:BBmod2:init} is numerically equal to $\mathbf{H}$~\eqref{eq:measurement-matrix-numeric}, but its even columns are in s$^{-1}$, instead of s.
    Notation is not altered for simplicity.
}
\begin{align}
\nonumber\\[-6mm]
	\mathbf{x}_{0|0}^{\twoDee} &= \mathbf{H}\T \mathbf{z}_0, & \mathbf{P}_{0|0}^{\twoDee} &= \mathbf{H}\T \mathbf{R} \mathbf{H} + \mathbf{V}, \label{eq:BBmod2:init}
 \\[-6mm]\nonumber
\end{align}
where $\mathbf{V}$ is constructed from the initial covariance matrix associated with the velocities as 
\begin{align}
\nonumber\\[-6mm]
	\mathbf{V} &=
    \blkdiag{0, v_{ \xVelocityTwoDee } (\mathbf{z}_0), 0, v_{ \yVelocityTwoDee } (\mathbf{z}_0), 0, v_{ \widthVelocityTwoDee } (\mathbf{z}_0), 0, v_{ \heightVelocityTwoDee } (\mathbf{z}_0) },
    \\[-6mm]\nonumber
\end{align}
where the initial variances for the velocities corresponding to $\xTwoDee_0, \yTwoDee_0, \widthTwoDee_0$ and $\heightTwoDee_0$ were computed as 
\begin{subequations} \label{eqs:initial_variances:2D}
\begin{align}
    \nonumber\\[-6mm]
    v_{ \xVelocityTwoDee } (\mathbf{z}_0) \ = \ v_{ \yVelocityTwoDee } (\mathbf{z}_0) &\triangleq
        \left( \tfrac{ [\mathbf{z}_0]_4 }{ \meanScalar{\heightThreeDee} } \cdot \tfrac{ \dot{r}^{\mathrm{max}} }{3} \right)^2 , \label{eqs:initial_variances:2D:a} \\
    v_{ \widthVelocityTwoDee } (\mathbf{z}_0) \ = \ v_{ \heightVelocityTwoDee } (\mathbf{z}_0) &\triangleq 
        \left( \tfrac{ [\mathbf{z}_0]_4 }{ \meanScalar{\heightThreeDee} } \cdot \tfrac{ \dot{s}^{\mathrm{max}} }{3} \right)^2 , \label{eqs:initial_variances:2D:b}
        \\[-6mm]\nonumber
\end{align}
\end{subequations}
where $[\, \cdot\, ]_{\ell}$ denotes the $\ell$-th element of the input vector (here the measured height), 
$\meanScalar{\heightThreeDee} = 1.65$ m is assumed mean pedestrian height in 3D, 
$\dot{r}^{\mathrm{max}} = 3$ m/s is assumed maximum pedestrian speed in 3D,
and $\dot{s}^{\mathrm{max}} = 0.3$ m/s is the assumed maximum temporal change rate of the pedestrian extent in 3D.

Equation~\eqref{eqs:initial_variances:2D:a} stems from transforming the corresponding standard deviation in 3D into the image plane based on:
\begin{itemize}
    \item Assuming the relation between the maximum and standard deviation of the pedestrian speed in 3D is $\dot{r}^{\mathrm{max}} = 3 \sigma_{\dot{r}}$.
    \item The depth can be estimated\footnote{
        The dept is estimated using~\eqref{eq:lineSegment:3DtoFocal} with $s_{\mathrm{O}}$ being the pedestrian height, substituted with the assumed mean pedestrian height $\meanScalar{\heightThreeDee}$.
        Using~\eqref{eq:FocalToImage:others}, $s_{\mathrm{I}}$ is then replaced with the measured height.
        Note that the height is preferable to the width for the depth estimation, as the width of a pedestrian, including its hands, may change significantly due to hand swinging.
    } as $\zThreeDee_{0|0} = \tfrac{ f \cdot \meanScalar{\heightThreeDee} }{ \px \cdot [\mathbf{z}_0]_4 }$.
    \item Applying~\eqref{eq:lineSegment:3DtoFocal} and~\eqref{eq:FocalToImage:others}.
\end{itemize}
Instead of using pedestrian speed, equation~\eqref{eqs:initial_variances:2D:b} follows by using the temporal change rate of the extent, i.e., $\dot{s}^{\mathrm{max}} = 3 \sigma_{\dot{s}}$.

\subsection{Bag-of-Tricks-Based  Filter}\label{sec:BoT-SORT}
The filter presented in this Section resulted from layering \emph{adjustments} to the Kalman filter originally used by the well-known SORT visual tracker~\cite{SORT:2016}. 

A notable adjustment was introduced in DeepSORT~\cite{DeepSORT:2017}, which suggested that the covariance matrices $\mathbf{Q}$ and $\mathbf{R}$, appearing in the Kalman filter should be functions of some estimated state or measurement elements (see their code).
Presumably, this should mimic the perspective projection effects.
\emph{Small} objects presumably in the background are given \emph{small} $\mathbf{Q}$ and $\mathbf{R}$, and vice-versa.
The resulting filter, however, is no longer a Kalman filter.
The final adjustments introduced in BoT-SORT include changing the original state vector\footnote{
    The original presentation from BoT-SORT paper~\cite{BoT-SORT:2022} and the corresponding code was revisited, leading to alternative notation in this paper.
}
to be 
\begin{align}
\\[-6mm]\nonumber
	\mathbf{x}_k^{\BoT} = [\xTwoDee_k \ms T\xVelocityTwoDee_k\ms \yTwoDee_k\ms T\yVelocityTwoDee_k\ms \widthTwoDee_k\ms T\widthVelocityTwoDee_k\ms \heightTwoDee_k\ms T\heightVelocityTwoDee_k]\T, \label{eq:BoTstate:variables}
 \\[-6mm]\nonumber
\end{align}
thus all units of the state variables in $\mathbf{x}_k^{\BoT}$~\eqref{eq:BoTstate:variables} are pixels.

The filter is initialized using the measurement $\mathbf{z}_0$ with mean $\mathbf{x}_{0|0}^{\BoT} = \mathbf{H}\T \mathbf{z}_{0}$ similarly\footnote{
    The matrix $\mathbf{H}$ used in the BoT-SORT filter is numerically equal to $\mathbf{H}$~\eqref{eq:measurement-matrix-numeric}. Its even columns are unitless, instead of being in seconds.
} to~\eqref{eq:BBmod2:init}, with initial covariance
\begin{align}
\\[-6mm]\nonumber
    \!\!\! \mathbf{P}_{0|0}^{\BoT} \!\! &= \!
    \mathcal{E}\left(
            [\mathbf{z}_0]_3, [\mathbf{z}_0]_4
            \right)
        \otimes
        \blkdiag{ 2^2 \zeta_{r}^2, 10^2 \zeta_{\dot{r}}^2 },
        \\[-6mm]\nonumber
\end{align}
where $\otimes$ denotes Kronecker product, $\zeta_{r} = \tfrac{1}{20}$ and $\zeta_{\dot{r}} = \tfrac{1}{160}$ are design parameters,
and $\mathcal{E}:\mathbb{R}^2\rightarrow \mathbb{R}^{4 \times 4}$ is defined as 
\begin{align}
\\[-6mm]\nonumber
    \mathcal{E} \left(
            \widthTwoDee, \heightTwoDee
    \right) = 
    \blkdiag{ \widthTwoDee^2, \heightTwoDee^2, \widthTwoDee^2, \heightTwoDee^2 }.
    \\[-6mm]\nonumber
\end{align}

The filtering recursion for $k=1,2,\dots,K$, is given by 
\begin{subequations}
\begin{align}
	\mathbf{x}_{k|k\!-\!1}^{\BoT} &= \mathbf{F}^{\BoT} \mathbf{x}_{k\!-\!1|k\!-\!1}^{\BoT} , \\
 \mathbf{P}_{k|k\!-\!1}^{\BoT} &= \mathbf{F}^{\BoT} \mathbf{P}_{k\!-\!1|k\!-\!1}^{\BoT} \left(\mathbf{F}^{\BoT}\right)\T + \mathbf{Q}_{k\!-\!1}^{\BoT} ,
\\[-7mm]\nonumber
        \end{align}
 \vspace*{-1mm}
 \end{subequations}
\vspace{-0.4cm}
\begin{subequations}\label{eqs:BoT:filter}
	\begin{align}
		\mathbf{x}_{k|k}^{\BoT} \! &= \! \mathbf{x}_{k|k\!-\!1}^{\BoT} + \mathbf{K}_k^{\BoT} ( \mathbf{z}_k \! - \! \mathbf{H}\mathbf{x}_{k|k\!-\!1}^{\BoT} ), \\
        \mathbf{P}_{k|k}^{\BoT} \! &= \! \mathbf{P}_{k|k\!-\!1}^{\BoT} \! - \! \mathbf{K}_k^{\BoT}  ( \mathbf{H} \mathbf{P}_{k|k\!-\!1}^{\BoT} \mathbf{H}\T \!\! + \! \mathbf{R}_k^{\BoT} ) \left( \mathbf{K}_k^{\BoT} \right)\T \!\!\!\! , \\
		\mathbf{K}_k^{\BoT} \! &= \! \mathbf{P}_{k|k\!-\!1}^{\BoT} \mathbf{H}\T ( \mathbf{H} \mathbf{P}_{k|k\!-\!1}^{\BoT} \mathbf{H}\T + \mathbf{R}_k^{\BoT} )^{-1} ,
	\end{align}
 \\[-5mm]\nonumber
\end{subequations}
where the custom matrices are given by 
\begin{align}
\\[-6mm]\nonumber
	\mathbf{F}^{\BoT} &= \blkdiag{
			\left[\begin{smallmatrix}
				1 & 1 \\ 0 & 1
			\end{smallmatrix}\right] ,
			\left[\begin{smallmatrix}
				1 & 1 \\ 0 & 1
			\end{smallmatrix}\right] ,
			\left[\begin{smallmatrix}
				1 & 1 \\ 0 & 1
			\end{smallmatrix}\right] ,
			\left[\begin{smallmatrix}
				1 & 1 \\ 0 & 1
			\end{smallmatrix}\right] }.
\\
	\mathbf{Q}_{k\!-\!1}^{\BoT} &=
        \mathcal{E}\left(
            \widthTwoDee_{k\!-\!1|k\!-\!1}^{\BoT}, \heightTwoDee_{k\!-\!1|k\!-\!1}^{\BoT}
            \right)
        \otimes
        \blkdiag{ \zeta_{r}^2, \zeta_{\dot{r}}^2 },
\\
	\mathbf{R}_{k}^{\BoT} &= 
        \mathcal{E}\left(
            \widthTwoDee_{k|k\!-\!1}^{\BoT}, \heightTwoDee_{k|k\!-\!1}^{\BoT}
        \right) \cdot \zeta_{r}^2.
        \\[-7mm]\nonumber
\end{align}

Notice that the 2D-Model-Based Filter~\ref{sec:twoDeeFilter} completely neglects any 3D modeling, while the Bag-of-Tricks-Based Filter~\ref{sec:BoT-SORT} takes it heuristically into account up to some level.
None of these filters produce estimates in 3D. 
These issues are tackled in the following Section by modeling directly in 3D.

\section{Proposed Model-Based Nonlinear Filter in 3D}\label{sec:novel_filter}
As discussed before, VTSs that model objects in 3D usually \emph{constrain} their motion to a common ground plane, which, however, requires additional knowledge of the plane.
This paper proposes a state-space model that allows \emph{unconstrained} object motion in 3D and a corresponding measurement model suited for monocular visual tracking-by-detection.
A nonlinear filter for this model is implemented using the UKF.

\subsection{Geometric Modeling in 3D}
This paper follows Fig.~\ref{fig:camera_geometry_2} to define a \emph{planar} BB in 3D.
Note that VTSs modeling three-dimensional blocks, however, use additional sensors~\cite{KITTI-dataset:2012} or assume ground plane motion~\cite{MonocularMTT3dOcclu:2013}. 

\newtheorem{definition}{Definition}
\begin{definition}[Planar BB in 3D]
    The \emph{planar 3D BB} is a perfectly aligned rectangle with the focal plane, as shown in Fig.~\ref{fig:camera_geometry_2}.
    All the vertices and edges of the planar 3D BB have the same depth in front of the camera.
    \hfill$\square$
\end{definition}
\vspace{0.1cm}

For a planar 3D BB, the state at time-step $k$ is defined as
\begin{align}
    \mathbf{x}_k^{\threeDee} = [ \xThreeDee_k\ \xVelocityThreeDee_k\ \yThreeDee_k\ \yVelocityThreeDee_k\ \zThreeDee_k\ \zVelocityThreeDee_k\ \widthThreeDee_k\ \heightThreeDee_k ]\T,\label{eq:state-3D-pom}
\end{align}
where the position $\xThreeDee_k$, $\yThreeDee_k$, $\zThreeDee_k$ of the lower-bottom center of the box and its width $\widthThreeDee_k$ and height $\heightThreeDee_k$ are in meters, while the velocities $\xVelocityThreeDee_k$, $\yVelocityThreeDee_k$, $\zVelocityThreeDee_k$ are in meters per second.
An example of such a 3D BB state is given in Fig.~\ref{fig:camera_geometry_2}, in which case position is $\mathbf{r}_{\mathrm{C,O}}=[\xThreeDee_k\ \yThreeDee_k\ \zThreeDee_k]\T$, the velocity is $\dot{\mathbf{r}}_{\mathrm{C,O}}=[\xVelocityThreeDee_k\ \yVelocityThreeDee_k\ \zVelocityThreeDee_k]\T$, the height is $s_{\mathrm{O}}=\heightThreeDee_k$, and the width is not denoted in Fig.~\ref{fig:camera_geometry_2}.

The dynamics~\eqref{eq:dynamics_nonlinear_general} of the 3D state is approximated as follows.
In accordance with standard motion modeling of physical objects~\cite{Bar-Shalom-et.al:2011}, an NCV model from Section~\ref{sec:NCV} is used to describe the position/velocity of the state $\mathbf{x}_k^{\threeDee}$~\eqref{eq:state-3D-pom}.
However, unlike the 2D case, the NCV model does not seem suitable for describing the extent. 
Therefore, this paper proposes to model the width and height using the following model.

\subsubsection{Auto-Regressive Model}
Consider an Ornstein-Uhlenbeck process $\dummyVar$ with a constant unconditional mean value $\Expect{\dummyVar(t)} = \meanScalar{}\in\mathbb{R}$ and variance $\var{\dummyVar(t)} = \sigma^2 >0$, which satisfies
\begin{align}
	\tfrac{d\dummyVar}{dt}(t) = - \tfrac{1}{ \tau } \dummyVar(t) + \tfrac{ \meanScalar{} }{\tau} + w(t),\ \var{w(t)} = \tfrac{2 \sigma^2}{\tau} \delta(0), \label{eq:rw_continuous_assumption}
\end{align}
where $\tau>0$ is a time-constant\footnote{
    Parameter $\tau$ can be specified as the elapsed time during which the process $\dummyVar(t)$, starting at $\dummyVar(0)$, changes its mean value to $\dummyVar(\tau) \approx 0.4 \dummyVar(0)+0.6\meanScalar{}$.
}, $w(t)$ is a zero-mean continuous white noise, and $\delta(t)$ is the Dirac delta function\footnote{
    Similarly to~\eqref{ncv_continuous_assumption}, a correct mathematical notation of~\eqref{eq:rw_continuous_assumption} would involve a stochastic differential equation and a Wiener process~\cite{Ornstein-Uhlenback:Maller:2009}.
}.

Discretization of \eqref{eq:rw_continuous_assumption} leads to~\cite{Ornstein-Uhlenback:Maller:2009}
\begin{align}
    \!\! \dummyVar_{k+1} = \alpha \dummyVar_k + \left(1\!-\!\alpha \right) \meanScalar{} + w_k,\ \var{w_k} = \sigma^2(1\!-\!\alpha^2) , 
    \label{eq:rw_discretized_directly}
\end{align}
where $\alpha = e^{ \sfrac{-\!T}{ \tau } }$.
The above discrete model is usually called the \emph{auto-regressive} (AR) model of order~1.

\subsubsection{State-Space Modeling in 3D}
Assuming, moreover, that the width $\widthThreeDee_k$ and height $\heightThreeDee_k$ obey the independent AR model, the state-space model has the form
\begin{subequations}\label{eq:3Dsystem}
\begin{align}
	\mathbf{x}_{k+1}^{\threeDee} &=
	\mathbf{F}^{\threeDee} \mathbf{x}_{k}^{\threeDee}
	+
    \mathbf{m}^{\threeDee}
	+
	\mathbf{w}_{k}^{\threeDee} , \label{eq:3Dsystem:state} \\
    \mathbf{z}_k &= \mathbf{H} \mathbf{p}(\mathbf{x}_k^{\threeDee}) + \mathbf{v}_k,
    \label{eq:3Dsystem:measurement}
\end{align}
\end{subequations}
where the dynamic equation~\eqref{eq:3Dsystem:state} is composed as follows.
The dynamic matrix and the additive term are
\begin{align}
    \mathbf{F}^{\threeDee} &= \blkdiag{ \mathbf{F}_{\mathrm{NCV}}, \mathbf{F}_{\mathrm{NCV}}, \mathbf{F}_{\mathrm{NCV}}, \alpha_{\widthThreeDee}, \alpha_{\heightThreeDee} }, \label{eq:3Dsystem:dynamicMatrix}\\
    \mathbf{m}^{\threeDee} &= 
    \left[ \mathbf{0}_{1 \times 6} \ms
        (1\!-\!\alpha_{\widthThreeDee})\!\cdot\!\meanScalar{\widthThreeDee} \ms
        (1\!-\!\alpha_{\heightThreeDee})\!\cdot\!\meanScalar{\heightThreeDee} \right]\T,
\end{align}
respectively, where $\mathbf{0}_{m \times n}$ is a zero matrix of dimensions $m \times n$, 
and where $\alpha_{\widthThreeDee} = e^{\sfrac{-\!T}{ \tau_{\widthThreeDee} }}$ and $\alpha_{\heightThreeDee} = e^{\sfrac{-\!T}{ \tau_{\heightThreeDee} }}$. The parameters of the width and height AR models are as follows.
The height of the planar 3D BB is modeled as that of a pedestrian, using the mean $\meanScalar{\heightThreeDee}=1.65$~m and time-constant $\tau_{\heightThreeDee} \approx 4$~s.
The width of the planar 3D BB is modeled with the mean $\meanScalar{\widthThreeDee} = 0.85$~m, and time-constant $\tau_{\widthThreeDee} \approx 0.4$~s to account for rough changes due to the hand swinging.
The matrix $\mathbf{Q}^{\threeDee} = \cov{\!\mathbf{w}_k^{\threeDee}\!}$ is
\begin{align}
	\mathbf{Q}^{\threeDee} = \blkdiag{ q_{\xVelocityThreeDee} \mathbf{T}, q_{\yVelocityThreeDee} \mathbf{T}, q_{\zVelocityThreeDee} \mathbf{T}, \sigma_{\widthThreeDee}^2(1\!-\!\alpha_{\widthThreeDee}^2), \sigma_{\heightThreeDee}^2(1\!-\!\alpha_{\heightThreeDee}^2) },
    \label{eq:3Dsystem:stateNoiseCov}
\end{align}
where the PSDs $q_{\xVelocityThreeDee} \! = \! q_{\yVelocityThreeDee} \! = \! q_{\zVelocityThreeDee} \! = \! 1$~m$^2$s$^{-3}$ are set for pedestrians 
~\cite[pp.~418]{Groves-NavigationSystems:2013}.
The standard deviations $\sigma_{\widthThreeDee} \! = \! \tfrac{0.45}{3}$~m and \linebreak$\sigma_{\heightThreeDee} \! = \! \tfrac{0.3}{3}$~m of the width and height, respectively, are selected based on their assumed relation to the maximal deviations from the means (not velocities as in Section~\ref{sec:BBmod2:initialization}), i.e., $\sigma_{\widthThreeDee}^{\mathrm{max}}=0.45$~m and $\sigma_{\heightThreeDee}^{\mathrm{max}}=0.3$~m, respectively.

The nonlinear measurement equation~\eqref{eq:3Dsystem:measurement} is described by the function $\mathbf{x}_k^{\twoDee} = \mathbf{p}(\mathbf{x}_k^{\threeDee})$, defined by\footnote{
    The function is constructed by stacking the perspective projection of individual state variables. According to the AR model, the variables $\dot{\widthThreeDee}_k$ and $ \dot{\heightThreeDee}_k$ are undefined and~\eqref{eq:lineSegment:3DtoFocal}-\eqref{eq:FocalToImage:others} is thus approximated with only $\dot{s}_{\mathrm{I}} = -\tfrac{f \zVelocityThreeDee}{\px \zThreeDee^2} s_{\mathrm{O}} $.
}
\begin{align}
\nonumber\\[-7mm]
	\mathbf{p}(\mathbf{x}_k^{\threeDee}) =
        \tfrac{f}{ \px \cdot \zThreeDee_k } 
        \left[\begin{smallmatrix}
			\xThreeDee_k \\
			\xVelocityThreeDee_k - \zVelocityThreeDee_k \cdot \xThreeDee_k / \zThreeDee_k \\
			\yThreeDee_k \\
			\yVelocityThreeDee_k - \zVelocityThreeDee_k \cdot \yThreeDee_k / \zThreeDee_k \\
			\widthThreeDee_k \\
			-\zVelocityThreeDee_k \cdot \widthThreeDee_k / \zThreeDee_k \\
			\heightThreeDee_k \\
			-\zVelocityThreeDee_k \cdot \heightThreeDee_k / \zThreeDee_k
        \end{smallmatrix}\right]
		+
        \left[\begin{smallmatrix}
			[\mathbf{r}_{\mathrm{I | F}}]_{1} \\
            0 \\
            [\mathbf{r}_{\mathrm{I | F}}]_{2} \\
			0 \\
			0 \\
			0 \\
			0 \\
			0
        \end{smallmatrix}\right]
        . \label{eq:T3d2d_state}
        \\[-6mm]\nonumber
\end{align}
Note that the measurement equation~\eqref{eq:3Dsystem:measurement} coincides with~\eqref{eq:system:measurement}, and thus the measurement noise covariance matrix is $\mathbf{R}$~\eqref{eq:state-noise-covariance-matrix:identification}.

\subsubsection{Physical Interpretation and Setup of Model Parameters}
The parameters specified above are suitable for describing pedestrians observed with a camera positioned in parallel with the ground or reasonably elevated\footnote{
    For the case of a top-view, where the height $\heightThreeDee_k$ can be assumed to coincide with the width $\widthThreeDee_k$, both processes can be modeled with the mean $\meanScalar{\widthThreeDee}=0.85$~m, standard deviation $\sigma_\widthThreeDee=\tfrac{0.45}{3}$~m, and time-constant $\tau_{\widthThreeDee}=0.4$.
}, i.e., if the height of the planar 3D BB state corresponds to the pedestrian's height well.
However, the physical modeling used in this paper allows for concise parameter selection, even for different situations.

In \cite{PedestrianStereoFiltering:2013}, the PSDs $q_{\xVelocityThreeDee}, q_{\yVelocityThreeDee},$ and $q_{\zVelocityThreeDee}$ were optimized for tracking pedestrians, suggesting values from 0.7 up to 0.77~m$^2$s$^{-3}$.
For vehicles, the PSDs can be selected as 10~m$^2$s$^{-3}$~\cite[pp.~418]{Groves-NavigationSystems:2013}.
For any case, the PSDs could also be specified according to the maximal acceleration $\ddot{r}^{\mathrm{max}}$ in m$\cdot$s$^{-2}$ as~\cite[pp.~44]{Bar-Shalom-et.al:2011}
\begin{align}
\nonumber\\[-7mm]
    q_{\xVelocityThreeDee} = q_{\yVelocityThreeDee} = q_{\zVelocityThreeDee} \approx \left( \ddot{r}^{\mathrm{max}} \right)^2 T,
    \\[-6mm]\nonumber
\end{align}
which is approximately equal to 1~m$^2$s$^{-3}$ for $T=\sfrac{1}{30}$~s and $\ddot{r}^{\mathrm{max}}=5.477$~m$\cdot$s$^{-2}$, i.e., if the maximal change of the object's speed over a second is $1.522$~km/h.
The speed of pedestrians for the MOT-17 dataset are also analyzed in \cite{MOT-15:2015}.

Parameters for the AR model could be easily stated for vehicles as well.
In \cite{PuttingObjectsInPerspective:2008}, 3D height statistical moments of both pedestrians and cars were estimated. 
Note that its selection should depend on the camera position and view angle.

\subsection{Model-Based Filter for \eqref{eq:3Dsystem}}
From~\eqref{eq:KF_update_mean}, note that the KF depends on the \emph{latest} measurement linearly.
If the state-space model is nonlinear and the KF cannot be applied, one can still define an optimal estimator with minimal MSE, subject to being linear w.r.t.~the \emph{latest} measurement, similarly to the KF.
One way to approximately implement\footnote{
    Assuming the depth $z_k$ and its uncertainty is such that the planar 3D BB is sufficiently far from the singularity at $z_k=0$. 
} the estimator is to use the \emph{unscented transform} (UT) leading to the \emph{unscented} Kalman filter (UKF)~\cite{Sarkka-Svensson:2023,square-root-UKF:2005}.

\subsubsection{Unscented Transform}
Consider a random vector $\mathbf{x}$ of dimension $n_{\mathbf{x}}$ with mean $\mathbf{m}_{\mathbf{x}}$ and covariance matrix $\mathbf{P}_{\mathbf{x}}$, and a random vector $\mathbf{y}=\mathbf{f}(\mathbf{x})$ resulting as a nonlinear transformation of $\mathbf{x}$.
The objective is to find an approximate mean $\mathbf{m}_{\mathbf{y}}$ and covariance matrix $\mathbf{P}_{\mathbf{y}}$ of $\mathbf{y}$.
Instead of approximating the nonlinear function, the UT uses deterministically chosen \emph{sigma-points}~\cite{Sarkka-Svensson:2023}.
The sigma-points are computed as\footnote{
    For simplicity, only a symmetric sigma-point set without additional scaling is considered in this paper.
    As a result, the formulation herein is equivalent to the spherical cubature approximation, but other formulations exist~\cite{Sarkka-Svensson:2023}.
}
\begin{align}
 \nonumber\\[-6mm]
    \!\!\mathcal{X}_i &=\begin{cases}
        \mathbf{m}_{\mathbf{x}} + \sqrt{ n_{\mathbf{x}} } \big( \sqrt{ \mathbf{P}_{\mathbf{x}} } \big)_{i},\ \text{for } i = 1,2,\dots,n_{\mathbf{x}}, \\
        \mathbf{m}_{\mathbf{x}} - \sqrt{ n_{\mathbf{x}} } \big( \sqrt{ \mathbf{P}_{\mathbf{x}} } \big)_{i},\ \text{for } i = n_{\mathbf{x}}+1,\dots,2\!\cdot\!n_{\mathbf{x}},
    \end{cases} \!\!\!\!\!
     \\[-6mm]\nonumber
\end{align}
where $(\,\cdot\,)_{i}$ denotes the $i$-th column of the input matrix, and $\sqrt{ \mathbf{P}_{\mathbf{x}} }$ is the matrix decomposition such that $\sqrt{ \mathbf{P}_{\mathbf{x}} } \left( \sqrt{ \mathbf{P}_{\mathbf{x}} } \right)\T = \mathbf{P}_{\mathbf{x}}$.
For $i=1,2,\dots,2\cdot n_{\mathbf{x}}$, the sigma-points are transformed via the nonlinear function to yield 
    $\mathcal{Y}_i = \mathbf{f}( \mathcal{X}_i )$.
The approximate mean and covariance matrix are then
\begin{subequations}
\begin{align}
 \nonumber\\[-6mm]
    \mathbf{m}_{\mathbf{y}} &= \textstyle \frac{1}{2\!\cdot\!n_{\mathbf{x}}} \sum_{i=1}^{2 \cdot n_{\mathbf{x}}} \mathcal{Y}_i,
    \label{eq:UKF:mean}\\
    \mathbf{P}_{\mathbf{y}} &= \mathbf{M}_{\mathbf{y}} \mathbf{M}_{\mathbf{y}}\T, \label{eq:UKF:cov}\\
    \mathbf{M}_{\mathbf{y}} &= \tfrac{1}{ \sqrt{ 2\!\cdot\!n_{\mathbf{x}} } } \big[ \mathcal{Y}_1\!-\!\mathbf{m}_{\mathbf{y}}\ms \mathcal{Y}_2\!-\!\mathbf{m}_{\mathbf{y}}\ms \dots \ms \mathcal{Y}_{2 n_{\mathbf{x}}}\!\!-\!\mathbf{m}_{\mathbf{y}} \big]. \label{eq:UKF:My}
    \\[-6mm] \nonumber
\end{align}
\end{subequations}
Moreover, the covariance $\cov{ \mathbf{x}, \mathbf{y} }$ can be approximated as
\begin{subequations}
\begin{align}
 \nonumber\\[-6mm]
    \mathbf{P}_{\mathbf{x},\mathbf{y}} &= \mathbf{M}_{\mathbf{x}} \mathbf{M}_{\mathbf{y}}\T, \\
    \mathbf{M}_{\mathbf{x}} &= \tfrac{1}{ \sqrt{ 2\!\cdot\!n_{\mathbf{x}} } } \big[ \mathcal{X}_1\!-\!\mathbf{m}_{\mathbf{x}}\ms \mathcal{X}_2\!-\!\mathbf{m}_{\mathbf{x}}\ms \dots \ms \mathcal{X}_{2 n_{\mathbf{x}}}\!\!-\!\mathbf{m}_{\mathbf{x}} \big]. \label{eq:UKF:Mx}
    \\[-6mm] \nonumber
\end{align}
\end{subequations}

\subsubsection{UKF Implementation of the Model-Based Filter}
The recursion starts at $k=0$ setting the initial filtering estimate as $\mathbf{x}_{0|0}^{\threeDee} = \Expect{ \mathbf{x}_0^{\threeDee} \,|\, \mathbf{z}_0 } $ and its error covariance matrix as $ \mathbf{P}_{0|0}^{\threeDee} = \cov{ \mathbf{x}_0^{\threeDee} \,|\, \mathbf{z}_0 }$.
For $k=1,2,\dots,K$, the prediction step is given by the Kalman filter prediction 
\begin{subequations}
	\begin{align}
 \nonumber\\[-6mm]
		\mathbf{x}_{k|k\!-\!1}^{\threeDee} &= \mathbf{F}^{\threeDee} \mathbf{x}_{k\!-\!1|k\!-\!1}^{\threeDee} + \mathbf{m}^{\threeDee}, \\
		\mathbf{P}_{k|k\!-\!1}^{\threeDee} &= \mathbf{F}^{\threeDee} \mathbf{P}_{k\!-\!1|k\!-\!1}^{\threeDee} \left(\mathbf{F}^{\threeDee}\right)\T + \mathbf{Q}^{\threeDee}.
   \\[-6mm]\nonumber
	\end{align}
\end{subequations}
The filtering step is given by the UKF update.
Using a numerically stable covariance matrix update based on the square-root UKF implementation~\cite{square-root-UKF:2005}, it is
\begin{subequations}
    \begin{align}
    \nonumber\\[-6mm]
        \mathbf{x}_{k|k}^{\threeDee} &= \mathbf{x}_{k|k\!-\!1}^{\threeDee} + \mathbf{K}_k^{\threeDee} ( \mathbf{z}_k - \mathbf{m}_{\mathbf{y}_k} ) , \\
        \mathbf{P}_{k|k}^{\threeDee} &= \big( \mathbf{M}_{\mathbf{x}_{k}^{\threeDee}} \!\! - \! \mathbf{K}_k^{\threeDee} \mathbf{M}_{\mathbf{y}_k} \! \big) \big( \cdot \big)\T \! + \mathbf{K}_k^{\threeDee}\mathbf{R} \! \left( \mathbf{K}_k^{\threeDee} \right)\T , \\
        \mathbf{K}_k^{\threeDee} &= \mathbf{M}_{\mathbf{x}_{k}^{\threeDee}} \mathbf{M}_{\mathbf{y}_k}\T \left( \mathbf{M}_{\mathbf{y}_k}\mathbf{M}_{\mathbf{y}_k}\T + \mathbf{R} \right)^{-1} ,
        \\[-6mm]\nonumber
    \end{align}
\end{subequations}
where $\mathbf{m}_{\mathbf{y}_k}$~\eqref{eq:UKF:mean}, $\mathbf{M}_{\mathbf{x}_{k}^{\threeDee}}$~\eqref{eq:UKF:Mx} and $\mathbf{M}_{\mathbf{y}_k}$~\eqref{eq:UKF:My}, are computed using the UT with function $\mathbf{y}_k = \mathbf{H} \mathbf{p}(\mathbf{x}_{k}^{\threeDee})$ and the mean $\mathbf{x}_{k|k\!-\!1}^{\threeDee}$ and covariance matrix $\mathbf{P}_{k|k\!-\!1}^{\threeDee}$ of 
$\mathbf{x}_{k}^{\threeDee}$.

If 2D BBs are needed in addition to the (predictive) estimates of the planar 3D BB states, one can directly output $\mathbf{m}_{\mathbf{y}_k}$~\eqref{eq:UKF:mean} and $\mathbf{P}_{\mathbf{y}_k}=\mathbf{M}_{\mathbf{y}_k} \mathbf{M}_{\mathbf{y}_k}\T$~\eqref{eq:UKF:cov}.
Note that the filtering 2D BB estimates are obtained analogically from $\mathbf{x}_{k|k}^{\threeDee}$ and $\mathbf{P}_{k|k}^{\threeDee}$.

\subsubsection{Initialization Using a Measurement}
For a fair comparison, the filter is initialized using the measurement $\mathbf{z}_0$ as
\begin{subequations}
\begin{align}
\nonumber\\[-6mm]
    \!\!\!\! \mathbf{x}_{0|0}^{\threeDee} \! &= \!
        \mathbf{C} \hat{\mathbf{r}}_{C,O}(\mathbf{z}_0) + 
        [\mathbf{0}_{1 \times 6} \ms \meanScalar{\widthThreeDee} \ms \meanScalar{\heightThreeDee}]\T,
    \label{eq:BBmod3:init:mean}\\
    \!\!\!\! \mathbf{P}_{0|0}^{\threeDee} \! &= \!
        \mathbf{C} \widehat{\mathbf{P}}_{C,O}(\mathbf{z}_0) \mathbf{C}\T \!\! +
        \blkdiag{
        \! 0, v_{\dot{r}}, 0, v_{\dot{r}} , 0, v_{\dot{r}} , \sigma_{\widthThreeDee}^2, \sigma_{\heightThreeDee}^2 \!
        } \! ,
    \label{eq:BBmod3:init:cov}
    \\[-6mm] \nonumber
\end{align}
\end{subequations}
where $\hat{\mathbf{r}}_{C,O}(\mathbf{z}_0)$ and $\widehat{\mathbf{P}}_{C,O}(\mathbf{z}_0)$ are the estimated mean and covariance of the initial position $[\xThreeDee_0\ \yThreeDee_0\ \zThreeDee_0]\T$ as described below, 
the auxiliary matrix $\mathbf{C} $ is numerically\footnote{
    The 2$^{\text{nd}}$, 4$^{\text{th}}$, and 6$^{\text{th}}$ rows of $\mathbf{C}$ are in s$^{-1}$, and the rest are unitless.
} equal to
\begin{align}
\nonumber\\[-6mm]
    \mathbf{C} =
    \begin{bmatrix}
            \mathbf{I}_{3} \otimes 
            \left[\begin{smallmatrix}
                1 \\ 0
            \end{smallmatrix}\right]
        \\
            \mathbf{0}_{2\times3}
    \end{bmatrix}
    \\[-6mm]\nonumber
\end{align}
with $\mathbf{I}_{n}$ being the identity matrix of dimension $n$,
and the initial variance $v_{\dot{r}}$ for the velocities 
is computed as
\begin{align}
\nonumber\\[-6mm]
    v_{\dot{r}} = 
    \left( \tfrac{ \dot{r}^{\mathrm{max}} }{3} \right)^2
    = 1\ \mathrm{m}^2\mathrm{s}^{-2}.
    \\[-6mm]\nonumber
\end{align}
To take the measurement noise into account, the variables $\hat{\mathbf{r}}_{C,O}(\mathbf{z}_0)=\mathbf{m}_\mathbf{r}$~\eqref{eq:UKF:mean} and $\widehat{\mathbf{P}}_{C,O}(\mathbf{z}_0)=\mathbf{P}_\mathbf{r}$~\eqref{eq:UKF:cov} are computed using the UT with function $\mathbf{r} = \mathbf{f} (\mathbf{x}; \mathbf{z}_0) $, specified as follows.
The random vector $\mathbf{x}$ is of dimension $n_{\mathbf{x}}=4$, has the mean $\mathbf{m}_{\mathbf{x}} = [\mathbf{0}_{1\times3}\ \meanScalar{\heightThreeDee}]\T$ and covariance matrix $\mathbf{P}_{\mathbf{x}} = \blkdiag{\mathbf{JRJ}\T, \sigma_{\heightThreeDee}^2 }$, where $\mathbf{J} \!=\!
    \left[\begin{smallmatrix}
        1 & 0 & 0 & 0 \\
        0 & 1 & 0 & 0 \\
        0 & 0 & 0 & 1 \\
    \end{smallmatrix}\right]$
is designed to omit the measurement noise corresponding to the width.
The function $\mathbf{f}(\,\cdot\,;\mathbf{z}_0):\mathbb{R}^{4}\rightarrow\mathbb{R}^{3}$ serves to compute an inverse of the noisy perspective projection for a 2D point $\left[ [\mathbf{z}_0]_1\ [\mathbf{z}_0]_2\right]\T$, i.e., of the first and third rows of~\eqref{eq:3Dsystem:measurement}, by estimating the depth\footnote{
    From Section~\ref{sec:BBmod2:initialization}, recall that using the height is preferable to the width for estimating the depth due to possible hand swinging of a boxed pedestrian.
} similarly to~\eqref{eqs:initial_variances:2D}, and it is given by
\begin{align}
\nonumber\\[-6mm]
    \mathbf{f}(\mathbf{x}; \mathbf{z}_0) = \frac{ [\mathbf{x}]_4 }{ [\mathbf{z}_0]_4 - [\mathbf{x}]_3 }
    \left[\begin{smallmatrix}
         [\mathbf{z}_0]_1 - [\mathbf{r}_{\mathrm{I,F}}]_1 - [\mathbf{x}]_1 \\
         [\mathbf{z}_0]_2 - [\mathbf{r}_{\mathrm{I,F}}]_2 - [\mathbf{x}]_2 \\
        \sfrac{f}{\px}
    \end{smallmatrix}\right].
    \\[-5mm]\nonumber
\end{align}
Note that the entire noisy perspective projection~\eqref{eq:3Dsystem:measurement} could be inverted and used to compute all entries of $\mathbf{x}_{0|0}^{\threeDee}$~\eqref{eq:BBmod3:init:mean} and $\mathbf{P}_{0|0}^{\threeDee}$~\eqref{eq:BBmod3:init:cov} based on $\mathbf{z}_0$, but it may not be viable.
For instance, the initial height and its covariance may be unreasonably large.
Instead, Equations~\eqref{eq:BBmod3:init:mean}-\eqref{eq:BBmod3:init:cov} are based on:
\begin{itemize}
    \item The initial position and its covariance are based on $\mathbf{z}_0$.
    \item The initial velocities are zero mean, and their variances are computed assuming the relation $\dot{r}^{\mathrm{max}} = 3 \sigma_{\dot{r}}$.
    \item The initial width and height and their variances are given by the modeled process parameters.
\end{itemize}

\section{Results With Real Data}
This section tests the filters using the training subset of the data from the publicly available dataset MOT-17 \cite{MOT-16:2016}.
The detections from Faster R-CNN VDN that are a part of the MOT-17 dataset are used \cite{MOT17-webpage:2023} and associated for individual annotations as in~\cite{KrKoSt:2023_FUSION}.
To exclude possible association errors, detections are also simulated.
Hand-selected parameters $\px\! =\! 10^{-6}$~m and $f\!=\!10^{-3}$~m were used.

\subsection{Retrieving of the 3D Information}
As only annotations in 2D are available in the studied data subset, a reliable evaluation for entire state vectors is not possible.
To arrive at indicative results, however, one can generate \emph{semi}-annotations in 3D of the form of a planar 3D BBs by computing the depth similarly to~\eqref{eqs:initial_variances:2D}, as $\zThreeDee_k \approx \tfrac{ f \cdot h^\mathrm{g} }{ \px \cdot [\mathbf{a}_k]_4 }$, where $h^\mathrm{g}$ in meters is the height of the given pedestrian \emph{guessed} based on observing images.
Applying~\eqref{eq:3Dto2dm}-\eqref{eq:FocalToImage:others}, the semi-annotations are
\begin{align}
    \mathbf{a}_k^{\threeDee} =
    \left[\begin{smallmatrix}
        \xThreeDee_k \\
        \yThreeDee_k \\
        \zThreeDee_k \\
        \widthThreeDee_k \\
        h^\mathrm{g}
    \end{smallmatrix}\right] \approx
     \frac{h^\mathrm{g}}{[\mathbf{a}_k]_{4}}
     \left[\begin{smallmatrix}
         [\mathbf{a}_k]_1 - [\mathbf{r}_{\mathrm{I,F}}]_1 \\
         [\mathbf{a}_k]_2 - [\mathbf{r}_{\mathrm{I,F}}]_2 \\
         \sfrac{f}{\px} \\
         [\mathbf{a}_k]_3 \\
         [\mathbf{a}_k]_4
     \end{smallmatrix}\right]
    \in \mathbb{R}^{5}.
\end{align}

\subsection{Evaluation Metrics}
To compare the filters, estimates corresponding to individual pedestrians are evaluated using two metrics.
Denote generally the ground-truth data (either annotation or 3D pseudo-annotation) with $\mathbf{x}_k\in\mathbb{R}^{n_{\mathbf{x}}}$ and the tuple of its filtering estimate and corresponding error covariance matrix of the $i$-th trial, $i=1,2,\dots,M$ with $\mathbf{x}_{k|k}^{(i)}, \mathbf{P}_{k|k}^{(i)}$.
Note that $M=1$ when using the Faster R-CNN detections. 

The \emph{root mean squared error} (RMSE) measures the absolute estimation error and is defined as
\begin{align}
\nonumber\\[-6mm]
    \textstyle
    \RMSE_k = \sqrt{ \frac{1}{M} \sum_{i=1}^{M} \left( \mathbf{x}_{k|k}^{(i)} - \mathbf{x}_k \right)\T \left( \mathbf{x}_{k|k}^{(i)} - \mathbf{x}_k \right) }.
    \label{eq:RMSE:def}
    \\[-6mm]\nonumber
\end{align}

The \emph{average normalized estimation error squared} (ANEES) takes the estimation error covariance into account as
\begin{align}
    \textstyle
    \!\!\!\! \ANEES_k \! = \! \frac{1}{M \cdot n_{\mathbf{x}}} \sum_{i=1}^{M} \!\! \left( \! \mathbf{x}_{k|k}^{(i)} \! - \! \mathbf{x}_k \! \right)\T \!\! \left( \! \mathbf{P}_{k|k}^{(i)} \! \right)^{-1} \!\! \left( \! \mathbf{x}_{k|k}^{(i)} \! - \! \mathbf{x}_k \! \right)\!. \!
    \label{eq:ANEES:ded}
\end{align}
ANEES equal to one indicates \emph{consistency}, i.e., the estimation error covariance matches the actual error.
However, ANEES being greater than one indicates \emph{overconfidence} of the estimate, i.e., the estimation covariance is on average too \emph{small} or \emph{optimistic} compared to the error, and ANEES being lower than one indicates \emph{under-confidence}, i.e., vice-versa.

\subsection{Results for Real and Simulated Detections}
To illustrate the performance of the presented filters, a couple of typical scenarios across the dataset are chosen and evaluated.
For both cases, $\gamma=1080$~px and $T=\sfrac{1}{30}$~s.
In addition, $200$ trials of simulated detections are generated using the available annotations and \eqref{eq:detection} with \eqref{eq:state-noise-covariance-matrix:identification}.

The legend of the following figures includes annotations or semi-annotations \annotationLGD{}; gray points \detectionsLGD{} indicating availability of detections; results with the 2D-Model-Based Filter (Section~\ref{sec:twoDeeFilter}) \twooDeeLGD{}, Bag-of-Tricks-Based Filter (Section~\ref{sec:BoT-SORT}) \BoTSORTLGD{}, and the proposed method \proposedLGD{}.
The dotted lines \camBoundariesLGD{} indicate edges of the camera's line of sight, and the stars \twooDeeStar{}, \BoTSORTStar{} and \proposedStar{} indicate medians computed over time-plots of the corresponding color.

\subsubsection{Sequence MOT17-02, pedestrian with id 2}\label{sec:results:MOT17-02:id2}
The scenario captures the pedestrian walking from the center to the edge of the image.
After being available for the first $1.9$ seconds, the detections are missing as the pedestrian reaches the edge.
The guessed height of the pedestrian is $h^\mathrm{g}=1.66$~m.

\begin{figure}[H]
\vspace{-6mm}
	\centering
    \subfloat{ \includegraphics[width=0.31\linewidth]{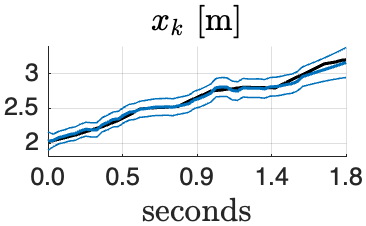} }
	\subfloat{ \includegraphics[width=0.31\linewidth]{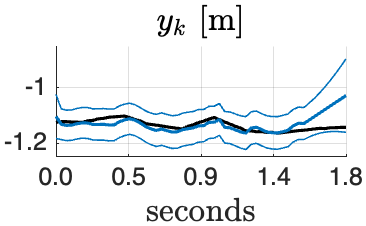} }
    \subfloat{ \includegraphics[width=0.31\linewidth]{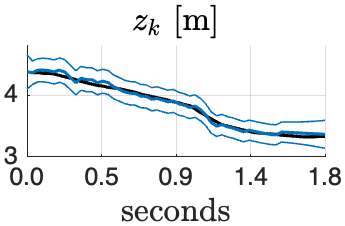} } \\
    \subfloat{ \includegraphics[width=0.31\linewidth]{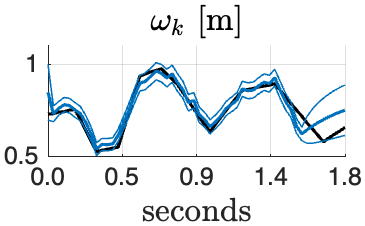} }
    \subfloat{ \includegraphics[width=0.31\linewidth]{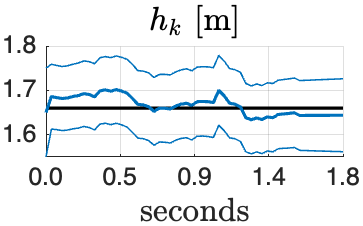} }
    \subfloat{ \includegraphics[width=0.31\linewidth]{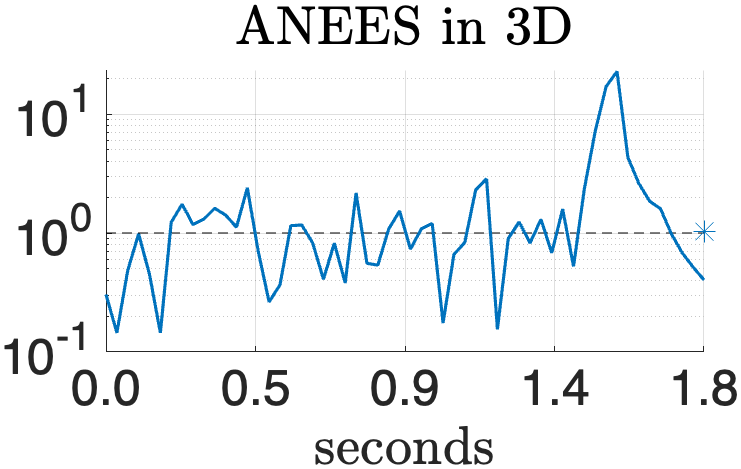} }
    \vspace{-2mm}
	\caption{Scenario \ref{sec:results:MOT17-02:id2}: estimated individual state variables of the proposed filter with standard deviation error intervals \proposedLGD{}, semi-annotations \annotationLGD{}, and ANEES with $\mathbf{x}_k=[\xThreeDee_k\ \yThreeDee_k\ \zThreeDee_k\ \widthThreeDee_k\ h^\mathrm{g}]\T$
 using Faster R-CNN detections.}\label{fig:MOT17-02_id2_fig3}
 \vspace{-8mm}
\end{figure}

\begin{figure}[H]
	\centering
    \subfloat{ \includegraphics[width=0.48\linewidth]{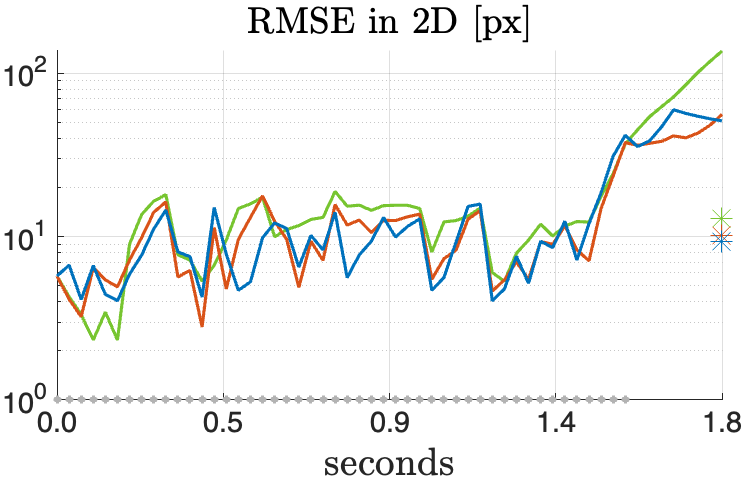} }
	\subfloat{ \includegraphics[width=0.48\linewidth]{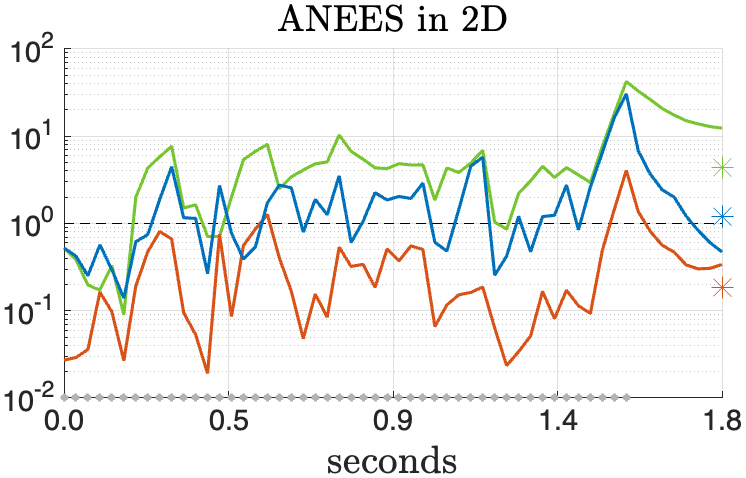} }
	\vspace{-2mm}
        \caption{Scenario \ref{sec:results:MOT17-02:id2}: RMSE and ANEES with $\mathbf{x}_k=[\xTwoDee_k\ \yTwoDee_k\ \widthTwoDee_k\ \heightTwoDee_k]\T$ and Faster R-CNN detections.}\label{fig:MOT17-02_id2_fig2}
\vspace{-6mm} 
\end{figure}

\begin{figure}[H]
\vspace{-2mm}
	\centering
	\subfloat{ \includegraphics[width=0.48\linewidth]{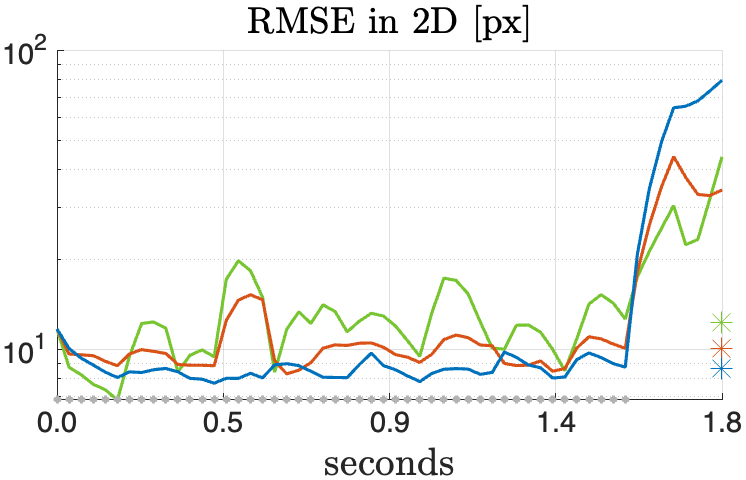} }
	\subfloat{ \includegraphics[width=0.48\linewidth]{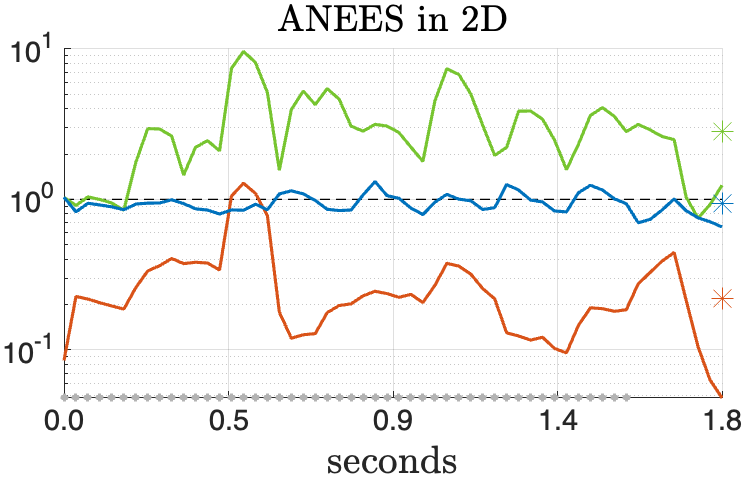} }
	\vspace{-2mm}
        \caption{Scenario \ref{sec:results:MOT17-02:id2}: RMSE and ANEES with $\mathbf{x}_k=[\xTwoDee_k\ \yTwoDee_k\ \widthTwoDee_k\ \heightTwoDee_k]\T$ and simulated detections.}\label{fig:MOT17-02_id2_SIMUL}
 \vspace{-2mm}
\end{figure}

Estimates from the proposed method based on true detections are presented in Fig.~\ref{fig:front} and Fig.~\ref{fig:MOT17-02_id2_fig3}.
The estimates in 2D and 3D (shown in $\xThreeDee–\zThreeDee$ view) with confidence ellipsoids are illustrated in Fig.~\ref{fig:front}, while the selected individual estimates and ANEES in 3D are plotted in Fig.~\ref{fig:MOT17-02_id2_fig3}.
RMSE and ANEES are shown for the estimates from all three presented methods based on real detections in  Fig.~\ref{fig:MOT17-02_id2_fig2} and simulated detections in Fig.~\ref{fig:MOT17-02_id2_SIMUL}.

In Fig.\ref{fig:front}, Fig.~\ref{fig:MOT17-02_id2_fig3}, Fig.~\ref{fig:MOT17-02_id2_fig2}, and Fig.~\ref{fig:MOT17-02_id2_SIMUL}, it can be seen that:
\begin{itemize}
    \item The estimates from the proposed method \proposedLGD{} track annotations in both 2D and 3D.
    \item The proposed method \proposedLGD{} provides consistent estimates as ANEES$_k$ fluctuates around one in both 2D and 3D. In contrast, the 2D-Model-Based Filter \twooDeeStar{} mostly provides optimistic estimates and the Bag-of-Tricks-Based Filter~\BoTSORTStar{} mostly provides pessimistic estimates. 
\end{itemize}
\begin{itemize}
    \item RMSE is mostly the lowest for the proposed method \proposedStar{} and the highest for the 2D-Model-Based Filter \twooDeeStar{}.
\end{itemize}

\subsubsection{Sequence MOT17-09, pedestrian with id 23}\label{sec:results:MOT17-09:id23}
The scenario captures the pedestrian walking from the image's left to the right edge.
In the second half of the scenario, detections are missing due to occlusion by other pedestrians for two seconds.
As the pedestrian reaches the edge, detections are missing again for $0.7$ seconds.
Here, $h^\mathrm{g}=1.71$~m.

\begin{figure}[H]
    \vspace{-8mm}
	\centering
	\subfloat[Filtering results in 2D.\label{fig:MOT17-09_id23_fig2:a}]{ \includegraphics[width=0.63\linewidth]{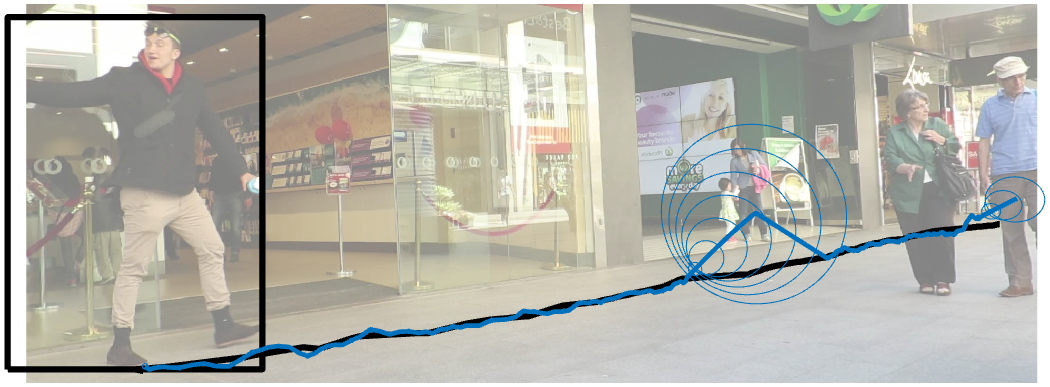} }
	\subfloat[Filtering results in 3D.\label{fig:MOT17-09_id23_fig2:b}]{ \includegraphics[width=0.33\linewidth]{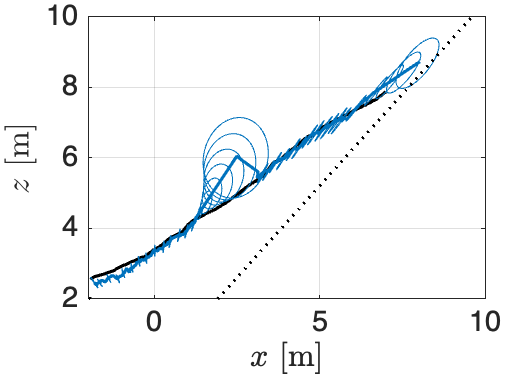} }
	\vspace{-2mm}
        \caption{Scenario \ref{sec:results:MOT17-09:id23}: Results of the proposed filter with the F-RCNN detector, including error covariance ellipses (plotted every 20 ms).}\label{fig:MOT17-09_id23_fig1}
        \vspace{-6mm}
\end{figure}

\begin{figure}[H]
	\centering
	\subfloat{ \includegraphics[width=0.48\linewidth]{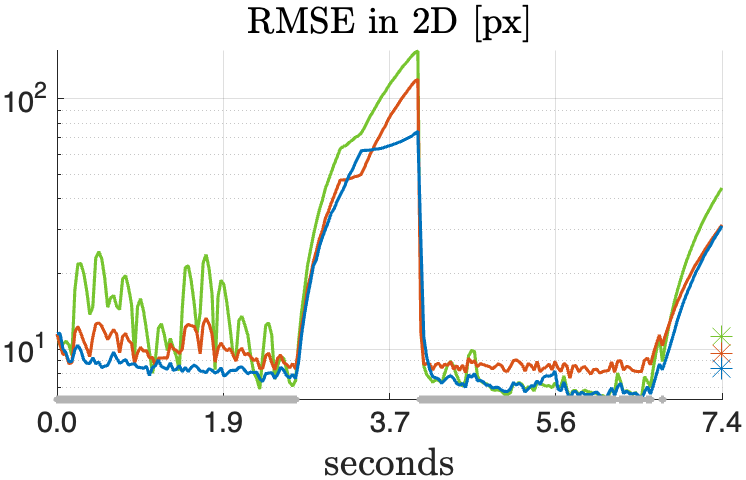} }
	\subfloat{ \includegraphics[width=0.48\linewidth]{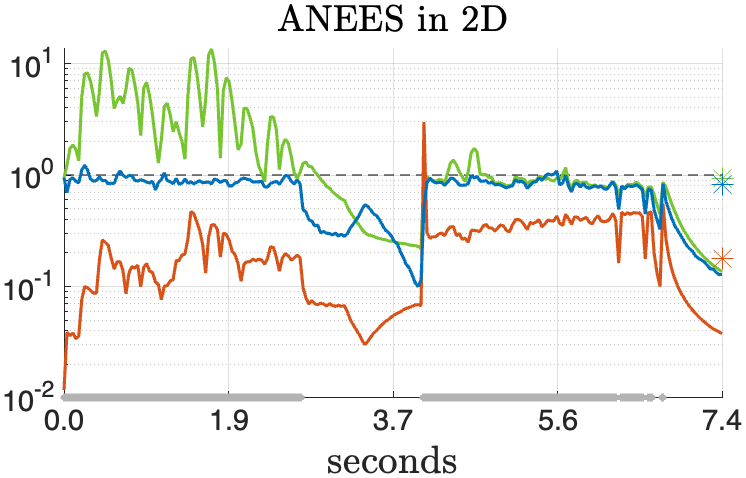} }
        \vspace{-2mm}
	\caption{Scenario \ref{sec:results:MOT17-09:id23}: RMSE and ANEES with $\mathbf{x}_k=[\xTwoDee_k\ \yTwoDee_k\ \widthTwoDee_k\ \heightTwoDee_k]\T$ and simulated detections.}\label{fig:MOT17-09_id23_SIMUL}
 \vspace{-2mm}
\end{figure}

Estimates from the proposed method based on true detections in 2D and 3D (shown in $\xThreeDee–\zThreeDee$ plane) with confidence ellipsoids are illustrated in Fig.~\ref{fig:MOT17-09_id23_fig1}.
RMSE and ANEES are shown for the estimates from all three presented methods based on real detections in  Fig.~\ref{fig:MOT17-02_id2_fig2} and simulated detections in Fig.~\ref{fig:MOT17-09_id23_SIMUL}.

\begin{figure}[H]
    \vspace{-2mm}
	\centering
    \subfloat{ \includegraphics[width=0.48\linewidth]{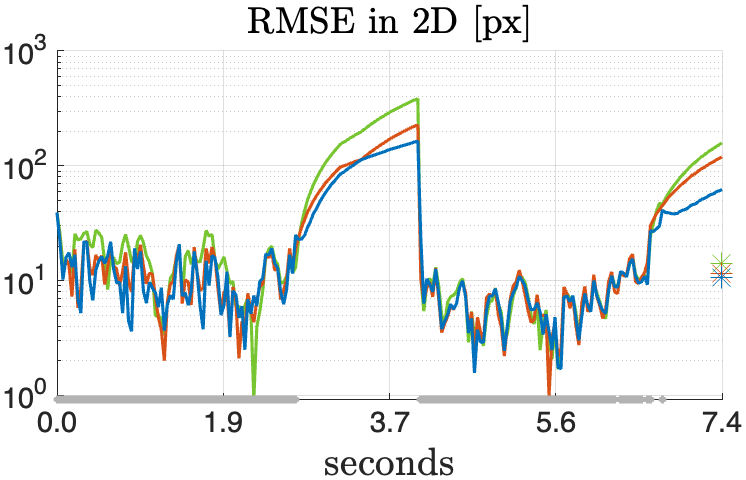} }
	\subfloat{ \includegraphics[width=0.48\linewidth]{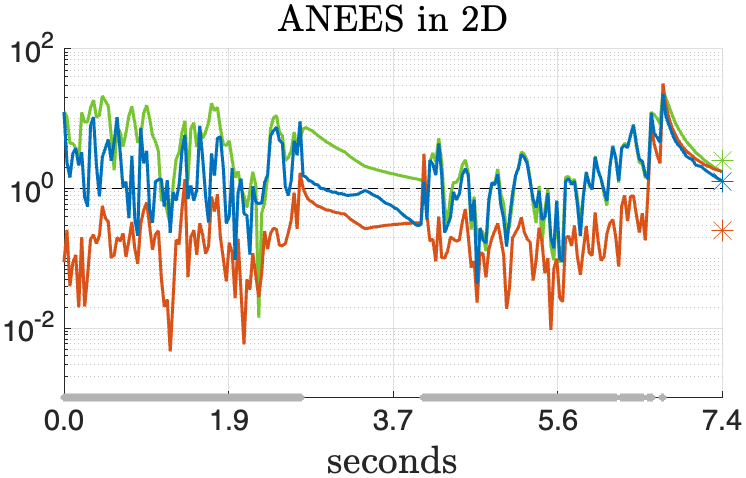} }
	\vspace{-2mm}
        \caption{Scenario \ref{sec:results:MOT17-09:id23}: RMSE and ANEES with $\mathbf{x}_k=[\xTwoDee_k\ \yTwoDee_k\ \widthTwoDee_k\ \heightTwoDee_k]\T$ and Faster R-CNN detections.}\label{fig:MOT17-09_id23_fig2}
        \vspace{-8mm}
\end{figure}

From Fig.\ref{fig:MOT17-09_id23_fig1}, Fig.~\ref{fig:MOT17-09_id23_fig2}, and Fig.~\ref{fig:MOT17-09_id23_SIMUL}, it can be concluded that:
\begin{itemize}
    \item The estimates from the proposed method \proposedLGD{} track annotations in both 2D and 3D.
\end{itemize}


\begin{itemize}
    \item
    The proposed method \proposedLGD{} provides consistent estimates in both 2D and 3D.
    The 2D-Model-Based Filter \twooDeeLGD{} provides mostly consistent estimates for real detections, but optimistic estimates for simulated detections in 2D.
    The Bag-of-Tricks-Based Filter \BoTSORTStar{} provides mostly pessimistic estimates for both real and simulated detections.
    \item In the second half of the scenario, the proposed filter \proposedLGD{} provides consistent estimates with the lowest RMSE for both real and simulated detections. 
    \item RMSE is mostly the lowest for the proposed method \proposedStar{}.\\[-2mm]
\end{itemize}

On average, the proposed method is about $10$ times slower than both baseline filters.
The Bag-of-Tricks-Based Filter is only about $10$\% slower than the 2D-Model-Based Filter.

\section{Conclusion}
The paper focused on visual tracking using a monocular camera and bounding boxes.
Often, traditional tracking approaches model the motion of the bounding box in 2D space, benefiting from the linear measurement equation.
To mimic the perspective projection, the state-of-the-art visual tracking algorithm BoT-SORT introduces heuristic approximations.
Tracking approaches working directly in 3D usually restrict the object motion to a known ground plane.

In this paper, a new model for object motion in 3D was constructed, in which object motion is not restricted to a common ground.
While the standard nearly constant velocity model models the bounding box location, the dynamics of width and height are represented by an autoregressive model.
Such a model facilitates convenient usage of the physical parameters of a tracked pedestrian.
While the model dynamics is linear, the measurement is a nonlinear function of the state due to the perspective projection.
Using the model in conjunction with the unscented Kalman filter leads to object's state estimates in 3D, which are consistent.
The estimate quality and consistency were illustrated numerically using the MOT-17 dataset.
Integration of the proposed filter into a VTS is left for future work.

As a future work, the 3D trajectories produced by the proposed algorithm could be used to estimate the ground plane.

\bibliographystyle{ieeetr}
\bibliography{references,literatura} 

\begin{thebibliography}{10}

\bibitem{Bar-Shalom-et.al:2011}
Y.~Bar-Shalom, P.~K. Willet, and X.~Tian, {\em Tracking and Data Fusion: A
  Handbook of Algorithms}.
\newblock YBS Publishing, 2011.

\bibitem{SonkaHlavacBoyle:ImageProcessing:2008}
M.~Sonka, V.~Hlavac, and R.~Boyle, {\em Image Processing, Analysis, and Machine
  Vision}.
\newblock Thomson Learning, third~ed., 2008.

\bibitem{SORT:2016}
A.~Bewley, Z.~Ge, L.~Ott, F.~Ramos, and B.~Upcroft, ``Simple online and
  realtime tracking,'' in {\em 2016 IEEE International Conference on Image
  Processing (ICIP)}, pp.~3464--3468, 2016.

\bibitem{BoT-SORT:2022}
N.~Aharon, R.~Orfaig, and B.-Z. Bobrovsky, ``{BoT-SORT}: {R}obust associations
  multi-pedestrian tracking.'' arXiv:2206.14651, 2022.

\bibitem{KrKoStDu:2023_FUSION}
J.~Krej{\v c}{\'\i}, O.~Kost, O.~Straka, and J.~Dun{\'\i}k, ``Bounding box
  dynamics in visual tracking: Modeling and noise covariance estimation,'' in
  {\em 2023 26th International Conf. on Information Fusion (FUSION)}, pp.~1--6,
  2023.

\bibitem{KrKoSt:2023_FUSION}
J.~Krej{\v c}{\'\i}, O.~Kost, and O.~Straka, ``Bounding box detection in visual
  tracking: Measurement model parameter estimation,'' in {\em 2023 26th
  International Conf. on Information Fusion (FUSION)}, pp.~1--8, 2023.

\bibitem{Surface3DMonoCamTracking:2022}
J.~Wang, W.~Choi, J.~Diaz, and C.~Trott, ``The 3d position estimation and
  tracking of a surface vehicle using a mono-camera and machine learning,''
  {\em Electronics}, vol.~11, no.~14, 2022.

\bibitem{MonocularMTT3dOcclu:2013}
C.~Wojek, S.~Walk, S.~Roth, K.~Schindler, and B.~Schiele, ``Monocular visual
  scene understanding: Understanding multi-object traffic scenes,'' {\em IEEE
  Tr. on Pat. Anal. and Mach. Int.}, vol.~35, no.~4, pp.~882--897, 2013.

\bibitem{UKF3D_StereoCam:2012}
T.~Junli and K.~Reinhard, ``Tracking of 2d or 3d irregular movement by a family
  of unscented {K}alman filters,'' {\em Journal of Inf. and Communication Conv.
  Engineering}, vol.~10, pp.~307--314, 09 2012.

\bibitem{KITTI-dataset:2012}
A.~Geiger, P.~Lenz, and R.~Urtasun, ``Are we ready for autonomous driving?
  {T}he {KITTI} vision benchmark suite,'' in {\em 2012 IEEE Conference on
  Computer Vision and Pattern Recognition}, pp.~3354--3361, 2012.

\bibitem{Mono-Camera3D_PMBM:2018}
S.~Scheidegger, J.~Benjaminsson, E.~Rosenberg, A.~Krishnan, and
  K.~Granstr{\"o}m, ``Mono-camera {3D} multi-object tracking using deep
  learning detections and {PMBM} filtering,'' in {\em 2018 IEEE Intelligent
  Vehicles Symposium (IV)}, pp.~433--440, 2018.

\bibitem{GroundPlaneUKF3d:2008}
M.~Meuter, U.~Iurgel, S.-B. Park, and A.~Kummert, ``The unscented {K}alman
  filter for pedestrian tracking from a moving host,'' in {\em 2008 IEEE
  Intelligent Vehicles Symposium}, pp.~37--42, 2008.

\bibitem{GroundPlaneEstim:2017}
T.~Liu, Y.~Liu, Z.~Tang, and J.-N. Hwang, ``Adaptive ground plane estimation
  for moving camera-based 3d object tracking,'' in {\em 2017 IEEE 19th Int.
  Workshop on Multimedia Signal Proc. (MMSP)}, pp.~1--6, 2017.

\bibitem{PedestrianKF_NCV:1997}
M.~Kohler, ``Using the {K}alman filter to track human interactive motion -
  modelling and initialization of the {K}alman filter for translational
  motion,'' tech. rep., Universit{\"a}t Dortmund, 1997.

\bibitem{KFpedestrianStereo_WH:2004}
M.~Bertozzi, A.~Broggi, A.~Fascioli, A.~Tibaldi, R.~Chapuis, and F.~Chausse,
  ``Pedestrian localization and tracking system with kalman filtering,'' in
  {\em IEEE Intelligent Vehicles Symp., 2004}, pp.~584--589, 2004.

\bibitem{MOT-16:2016}
A.~Milan, L.~Leal-Taixe, I.~Reid, S.~Roth, and K.~Schindler, ``{MOT16}: A
  benchmark for multi-object tracking,'' {\em arXiv:1603.00831}, 2016.

\bibitem{MOT17-webpage:2023}
A.~Milan, L.~Leal-Taix{\'e}, I.~Reid, S.~Roth, and K.~Schindler, ``website of
  the {Multiple Object Tracking Benchmark {MOT17}},'' {\em at
  \url{https://motchallenge.net/data/MOT17/}}, last checked 2024, March 14.

\bibitem{PuttingObjectsInPerspective:2008}
D.~Hoiem, A.~A. Efros, and M.~Hebert, ``Putting objects in perspective,'' {\em
  International Journal of Computer Vision}, vol.~80, no.~1, pp.~3--15, 2008.

\bibitem{MOT-15:2015}
L.~Leal-Taix\'{e}, A.~Milan, I.~Reid, S.~Roth, and K.~Schindler,
  ``{MOTC}hallenge 2015: Towards a benchmark for multi-target tracking,'' {\em
  arXiv:1504.01942}, 2015.

\bibitem{MOT-20:2020}
P.~Dendorfer, H.~Rezatofighi, A.~Milan, J.~Shi, D.~Cremers, I.~Reid, S.~Roth,
  K.~Schindler, and L.~Leal-Taixe, ``{MOT}20: {A} benchmark for multi object
  tracking in crowded scenes,'' {\em arXiv:2003.09003}, 2020.

\bibitem{DeepSORT:2017}
N.~Wojke, A.~Bewley, and D.~Paulus, ``Simple online and realtime tracking with
  a deep association metric.'' arXiv:1703.07402, 2017.

\bibitem{Ornstein-Uhlenback:Maller:2009}
R.~A. Maller, G.~M{\"u}ller, and A.~Szimayer, {\em Ornstein--Uhlenbeck
  Processes and Extensions}, pp.~421--437.
\newblock Springer Berlin Heidelberg, 2009.

\bibitem{Groves-NavigationSystems:2013}
P.~D. Groves, {\em {P}rinciples of {GNSS}, Inertial, and Multisensor Integrated
  Navigation Systems}.
\newblock Artech House, second~ed., 2013.

\bibitem{PedestrianStereoFiltering:2013}
N.~Schneider and D.~M. Gavrila, ``Pedestrian path prediction with recursive
  bayesian filters: A comparative study,'' in {\em Pattern Recognition}
  (J.~Weickert, M.~Hein, and B.~Schiele, eds.), (Berlin, Heidelberg),
  pp.~174--183, Springer Berlin Heidelberg, 2013.

\bibitem{Sarkka-Svensson:2023}
S.~S{\"a}rkk{\"a} and L.~Svensson, {\em {B}ayesian Filtering and Smoothing.
  Second Edition.}
\newblock Institute of Mathematical Statistics Textbooks, Cambridge University
  Press, 2023.

\bibitem{square-root-UKF:2005}
M.~Simandl and J.~Dun{\'\i}k, ``Sigma point gaussian sum filter design using
  square root unscented filters,'' {\em IFAC Proceedings Volumes}, vol.~38,
  no.~1, pp.~1000--1005, 2005.
\newblock 16th IFAC World Congress.

\end{thebibliography}

\end{document}